\documentclass[preprint,12pt]{elsarticle}

\usepackage{amssymb}
\usepackage{amsmath}
\usepackage{amsfonts}
\usepackage{bm} 
\usepackage{algorithm}
\usepackage[noend]{algpseudocode}
\usepackage{array}
\usepackage{textcomp}
\usepackage{url}
\usepackage{verbatim}
\usepackage{cite}
\usepackage{color}
\usepackage[inline]{enumitem}
\usepackage{upgreek}
\usepackage{booktabs}
\usepackage{ragged2e}
\usepackage[para,online,flushleft]{threeparttable}
\usepackage{mathtools}
\usepackage{subcaption}

\DeclareMathOperator*{\argmax}{arg\,max}

\newcommand{\cop}[1]{{#1}\allowbreak} 

\journal{Robotics and Autonomous Systems}

\begin{document}

\begin{frontmatter}



\title{Omnidirectional Sensor Placement: \\ A Large-Scale Computational Study and Novel \\ Hybrid Accelerated-Refinement Heuristics}
\tnotetext[note1]{This work was co-funded by the European Union under the project Robotics and advanced industrial production (reg.~no.~\texttt{CZ.02.01.01/\allowbreak{}00/\allowbreak{}22\_008/\allowbreak{}0004590}) and by the Grant Agency of the Czech Technical University in Prague, grant no.~\texttt{SGS23/\allowbreak{}175/\allowbreak{}OHK3/\allowbreak{}3T/\allowbreak{}13}.}
\author[label1,label2]{Jan Mikula\corref{cor1}}
\ead{jan.mikula@cvut.cz}
\author[label1]{Miroslav Kulich}
\ead{miroslav.kulich@cvut.cz}

\cortext[cor1]{Corresponding author}

\affiliation[label1]{organization={Czech Institute of Informatics, Robotics and Cybernetics, Czech Technical University in Prague},
            addressline={Jugoslavskych partyzanu 1580/3},
            city={Prague~6},
            postcode={160\,00},
            country={Czech Republic}}

\affiliation[label2]{organization={Department of Cybernetics, Faculty of Electrical Engineering, Czech Technical University in Prague},
            addressline={Karlovo namesti 293/13},
            city={Prague~2},
            postcode={121\,35},
            country={Czech Republic}}

\begin{abstract}
    This paper studies the \emph{omnidirectional sensor-placement problem} (OSPP), which involves placing static sensors in a continuous 2D environment to achieve a user-defined coverage requirement while minimizing sensor count.
    The problem is motivated by applications in mobile robotics, particularly for optimizing visibility-based route planning tasks such as environment inspection, target search, and region patrolling.
    We focus on \emph{omnidirectional visibility models}, which eliminate sensor orientation constraints while remaining relevant to real-world sensing technologies like LiDAR, 360-degree cameras, and multi-sensor arrays.
    Three key models are considered: \emph{unlimited visibility}, \emph{limited-range visibility} to reflect physical or application-specific constraints, and \emph{localization-uncertainty visibility} to account for sensor placement uncertainty in robotics.
    Our first contribution is a large-scale computational study comparing classical convex-partitioning and sampling-based heuristics for the OSPP, analyzing their trade-off between runtime efficiency and solution quality.
    Our second contribution is a new class of \emph{hybrid accelerated-refinement} (HAR) heuristics, which combine and refine outputs from multiple sensor-placement methods while incorporating preprocessing techniques to accelerate refinement.
    Results demonstrate that HAR heuristics significantly outperform traditional methods, achieving the lowest sensor counts and improving the runtime of sampling-based approaches.
    Additionally, we adapt a specific HAR heuristic to the localization-uncertainty visibility model, showing that it achieves the required coverage for small to moderate localization uncertainty.
    Future work may apply HAR to visibility-based route planning tasks or explore novel sensor-placement approaches to achieve formal coverage guarantees under uncertainty.
\end{abstract}

\begin{graphicalabstract}
    \begin{figure*}[b!]
        \centering
        \caption*{\textbf{Omnidirectional Visibility Models:}}
        \begin{subfigure}[t]{0.3\columnwidth}
            \centering
            \includegraphics[width=\linewidth]{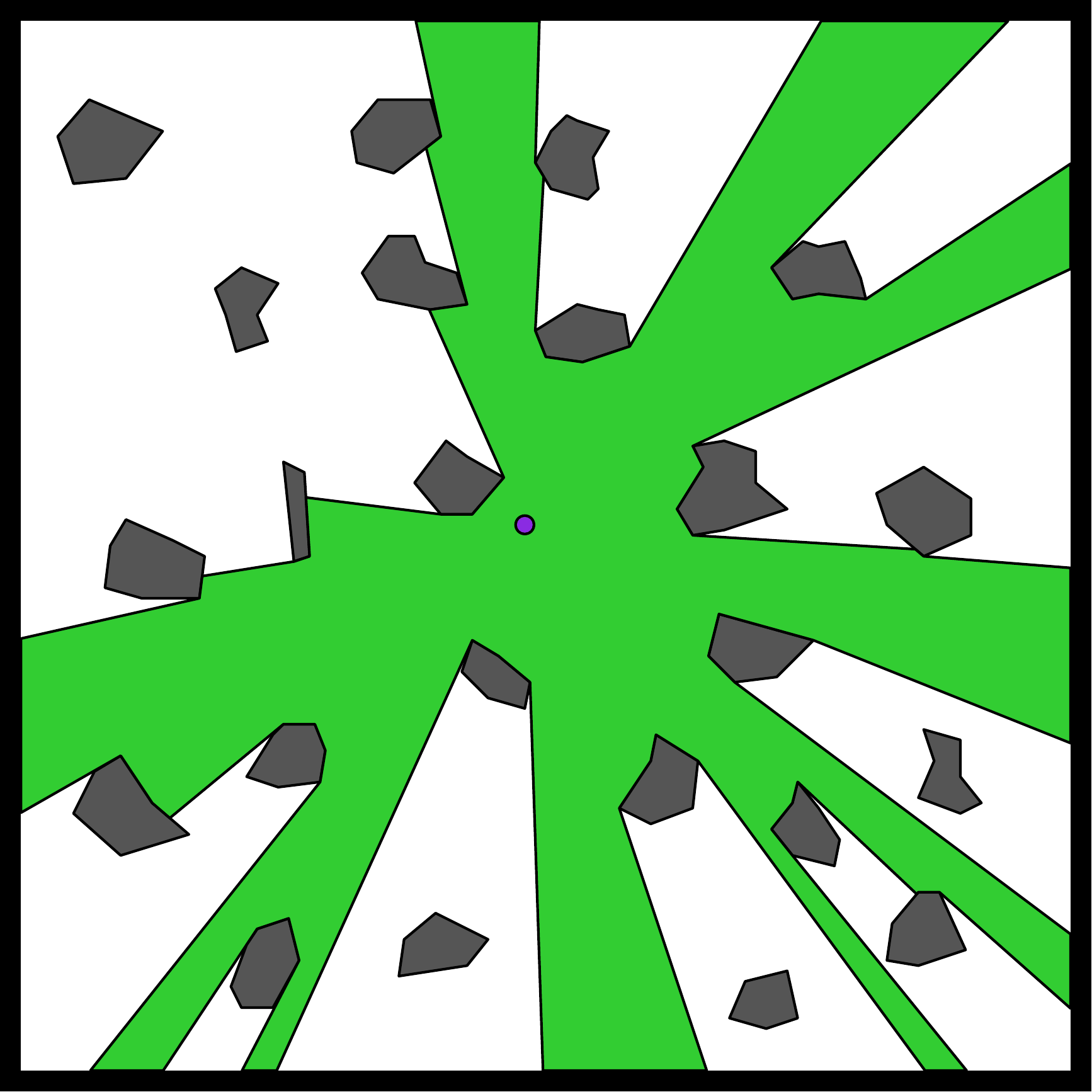}
            \caption*{Unlimited Visibility}
        \end{subfigure}
        \hfill
        \begin{subfigure}[t]{0.3\columnwidth}
            \centering
            \includegraphics[width=\linewidth]{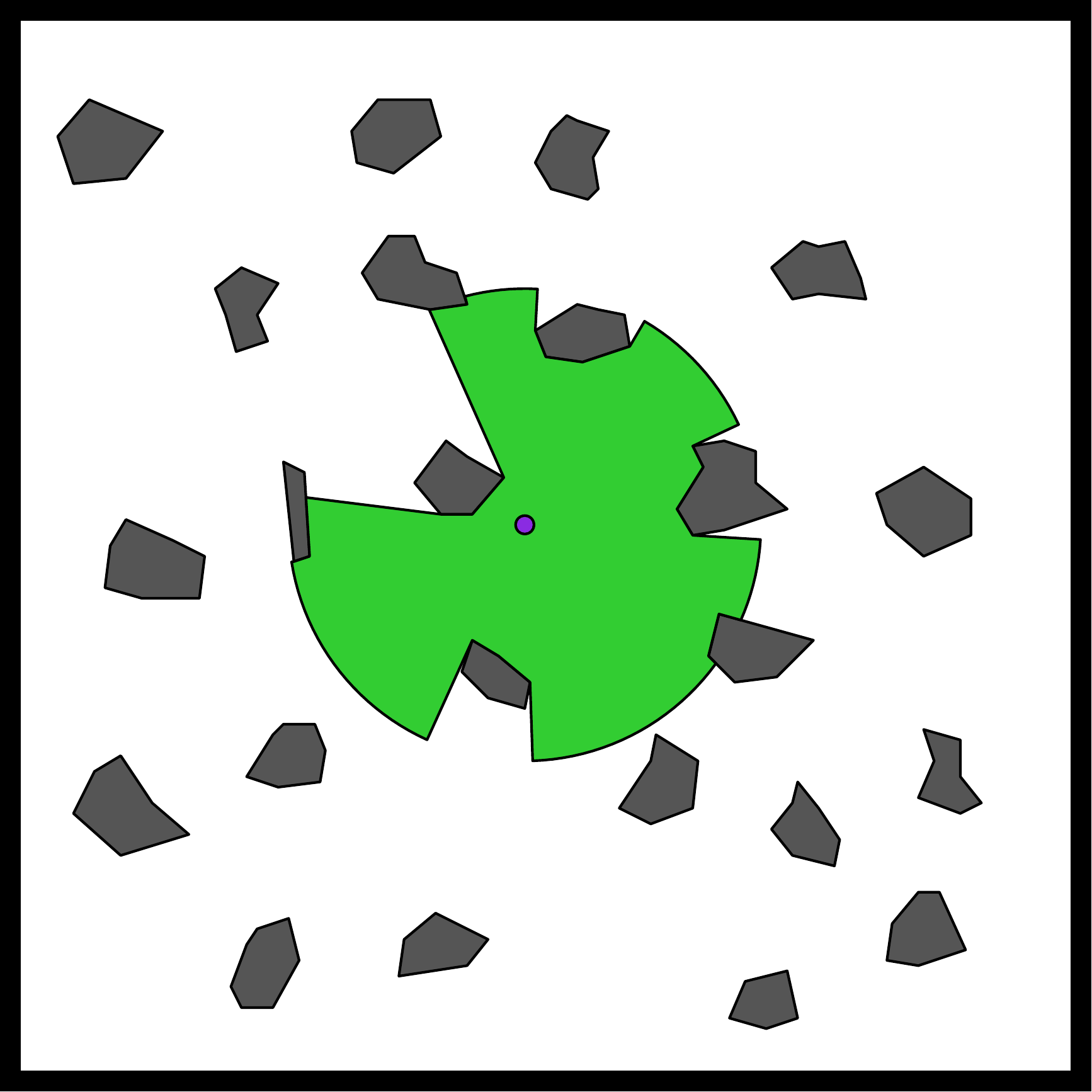}
            \caption*{Limited-Range Visibility}
        \end{subfigure}
        \hfill
        \begin{subfigure}[t]{0.3\columnwidth}
            \centering
            \includegraphics[width=\linewidth]{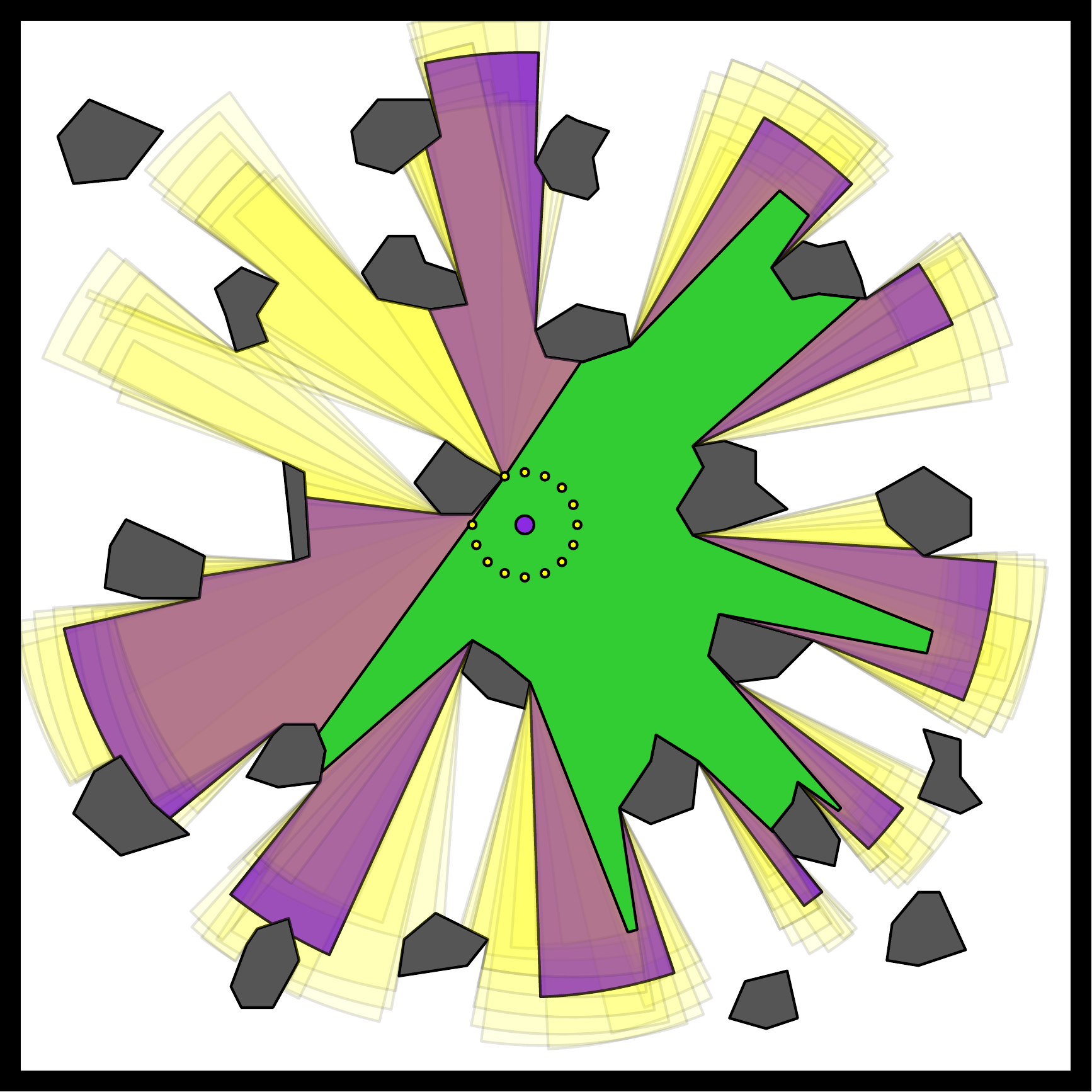}
            \caption*{Localization-Uncertainty Visibility}
        \end{subfigure}
        \\\vspace{1em}
        \caption*{\textbf{Principles of Hybrid Accelerated-Refinement Heuristics:}}
        \begin{subfigure}[t]{0.3\columnwidth}
            \centering
            \includegraphics[width=\linewidth]{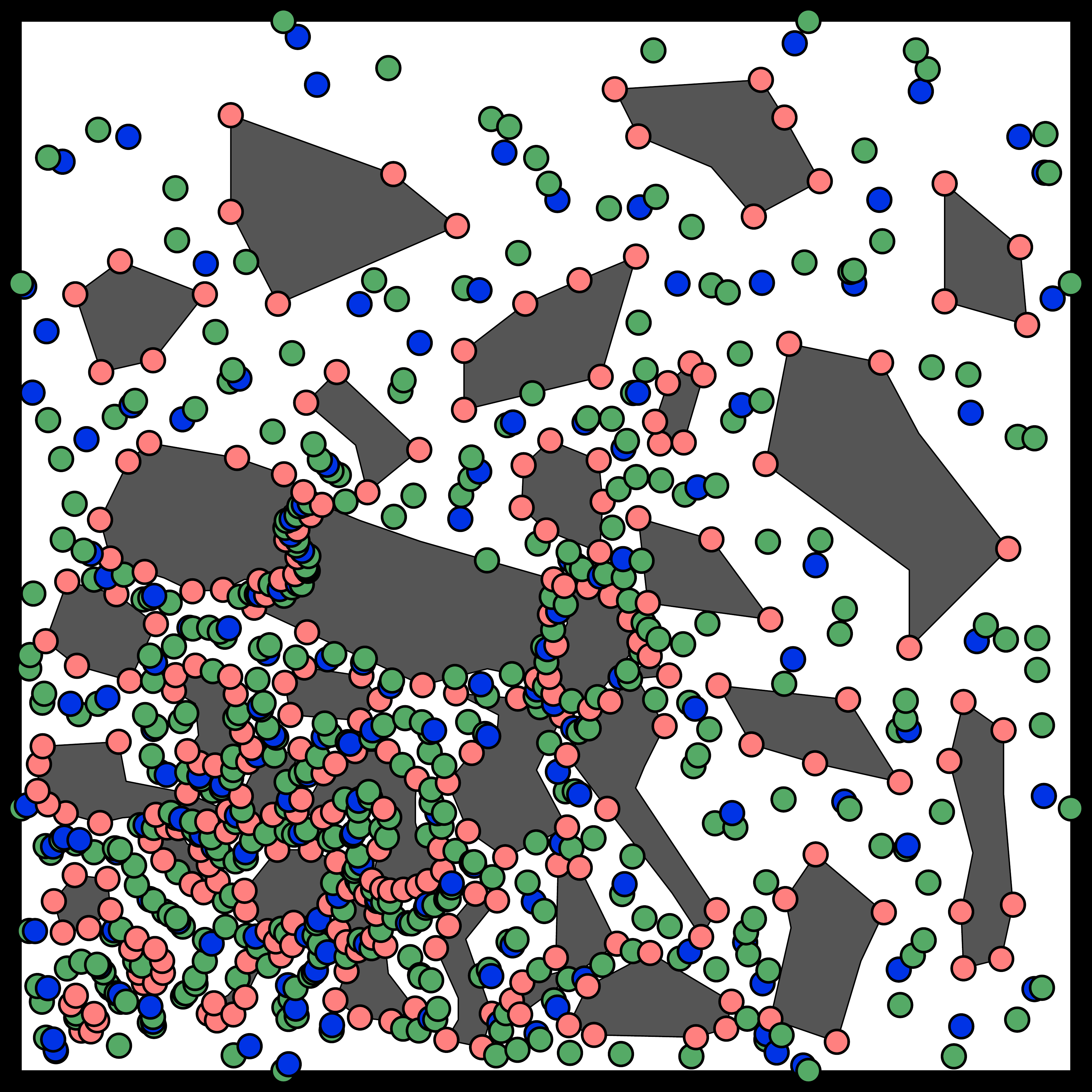}
            \caption*{Combine Multiple Guard Methods}
            \caption*{(Initial Count: 821)}
        \end{subfigure}
        \hfill
        \begin{subfigure}[t]{0.3\columnwidth}
            \centering
            \includegraphics[width=\linewidth]{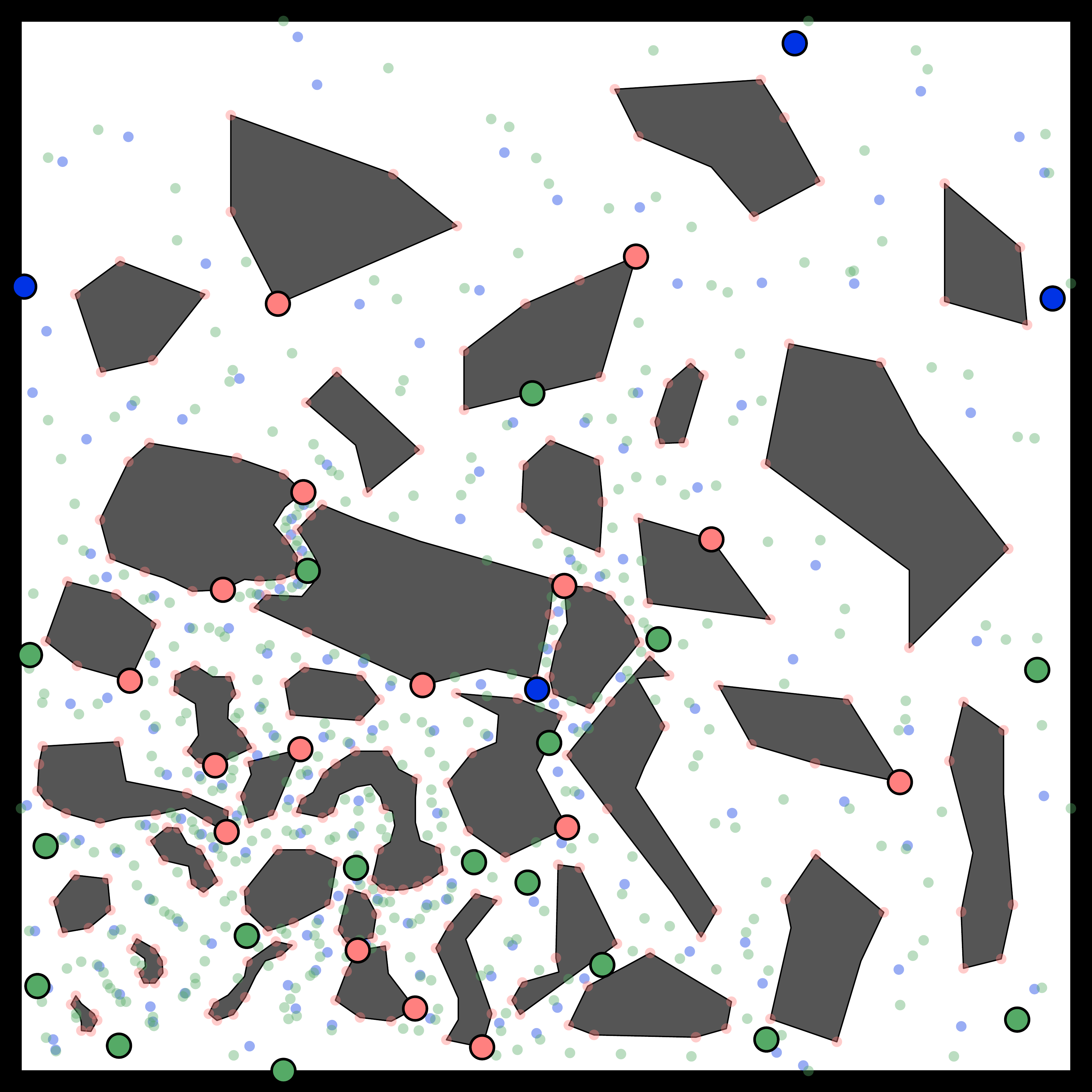}
            \caption*{Preprocess Coverage and Refine}
            \caption*{(Refined Count: 37)}
        \end{subfigure}
        \hfill
        \begin{subfigure}[t]{0.3\columnwidth}
            \centering
            \includegraphics[width=\linewidth]{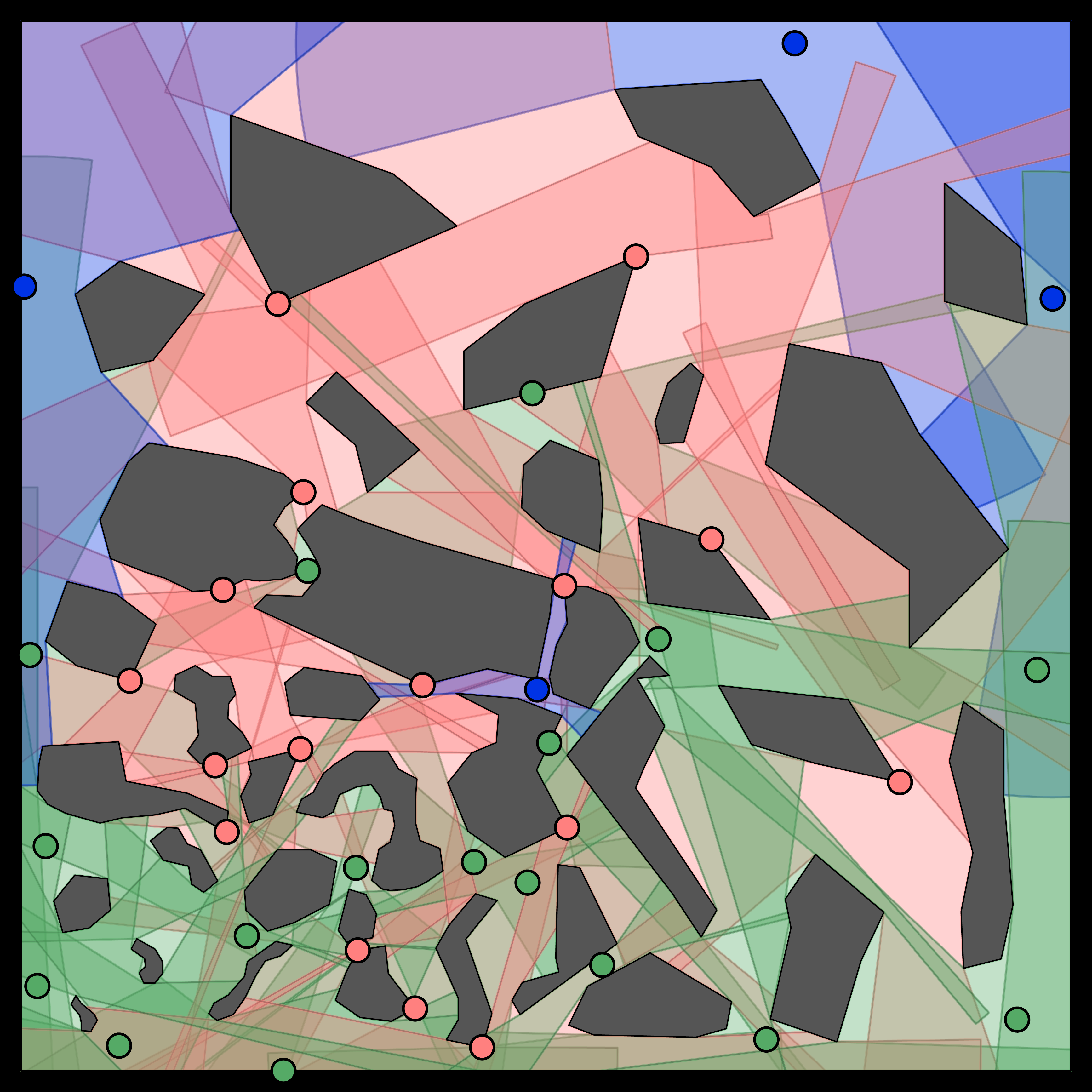}
            \caption*{Refined Coverage}
        \end{subfigure}
    \end{figure*}
    \setcounter{figure}{0} 
\end{graphicalabstract}

\begin{highlights}
    \item Studies the omnidirectional sensor-placement problem in continuous 2D environments
    \item Evaluates unlimited, limited-range, and localization-uncertainty visibility models
    \item Conducts a large-scale study of convex-partitioning and sampling heuristics
    \item Introduces hybrid accelerated-refinement heuristics, taking the best of both worlds
    \item Novel heuristics provide best trade-off between sensor count and runtime efficiency
\end{highlights}


\end{frontmatter}



\section{Introduction}
\label{sec:introduction}

This paper addresses a variant of the \emph{sensor-placement problem} (SPP), which involves strategically positioning static sensors in a known 2D continuous environment.
Typically, the objective is to achieve full coverage while minimizing the number of sensors, where coverage is determined by the union of visible regions from each sensor, as defined by a specified visibility model.
SPPs, including variants with additional constraints and modified objectives, have applications across diverse domains such as urban surveillance, environmental monitoring, smart agriculture, and robotics.
For instance, sensor placement is essential for transport network surveillance and urban monitoring~\citep{Murray2007,Kritter2019}, structural health monitoring of bridges and buildings~\citep{Ostachowicz2019}, and optimizing agricultural field coverage~\citep{An2015}.
In robotics, the SPP can be tailored to enhancing mobile robot navigation~\citep{Vitus2011} or assisting in planning efficient inspection routes~\citep{Faigl2011}.

In this paper, we study the SPP in the general context of mobile robotics, with a focus on optimizing visibility-based route planning tasks such as environment inspection, target search, and region patrolling.
A standard approach for these tasks follows a \emph{decoupling scheme}~\citep[used, e.g., in][]{Sarmiento2004,Packer2008,Faigl2011}, where the problem is divided into two subproblems: \emph{sensor placement} (addressed in this work) and \emph{route optimization}, which typically involves solving NP-hard combinatorial optimization problems depending on the task objectives, such as the \emph{traveling salesperson problem}~\citep{Faigl2011}, the \emph{traveling delivery person problem}~\citep{Mikula2022}, the \emph{graph search problem}~\citep{Kulich2017}, \emph{expected-time mobile search}~\citep{Sarmiento2004}, and \emph{vehicle routing problems}~\citep{Macharet2018}.
While we do not explicitly address a specific route planning problem, this context underpins our approach to the SPP, emphasizing \emph{minimum} sensor placement to reduce the complexity of route optimization.
Additionally, it motivates our focus on runtime efficiency and the ability to handle large, complex, continuous environments, ensuring applicability to real-world long-term robotic missions.

Our study focuses on \emph{omnidirectional visibility models}, which are independent of sensor orientation, reducing computational complexity and enabling efficient solutions in large-scale, complex environments while maintaining real-world applicability.
Many commonly used robotic sensor systems, such as LiDAR, 360-degree cameras, and multi-sensor arrays, naturally provide omnidirectional or approximately omnidirectional coverage, making them well-suited for the models we consider.
We evaluate three key visibility models: the \emph{classical unlimited visibility} model, the \emph{limited-range visibility} model, which accounts for physical constraints or application-specific requirements such as resolution limitations in camera-based object detection, and the \emph{localization-uncertainty visibility} model, motivated by mobile robotics scenarios where the sensor’s exact placement is uncertain due to control imprecision or localization errors.
By focusing on these models, we balance computational efficiency with practical relevance, ensuring applicability to real-world robotic sensing tasks.

We formulate the SPP with omnidirectional sensors as the \emph{omnidirectional} SPP (OSPP), which involves a continuous 2D environment with obstacles, a general omnidirectional visibility model, and a user-defined coverage requirement introduced to mitigate the diminishing returns of achieving full coverage in complex environments.
Given the problem's inherent complexity, we focus on \emph{heuristic solution methods}, which do not guarantee optimality in minimizing the number of sensors but offer the potential for high-quality practical solutions.
While several classical heuristics applicable to the OSPP exist, particularly in the \emph{convex-partitioning}~\citep{Kazazakis2002} and \emph{sampling-based}~\citep{Gonzalez-Banos1998} categories, a large-scale computational study comparing these methods and examining the trade-off between minimizing sensor count and ensuring runtime efficiency is still lacking.
Addressing this gap is our first contribution.

\begin{figure}[t]
    \centering
    \begin{subfigure}[t]{0.155\columnwidth}
        \centering
        \includegraphics[width=\linewidth]{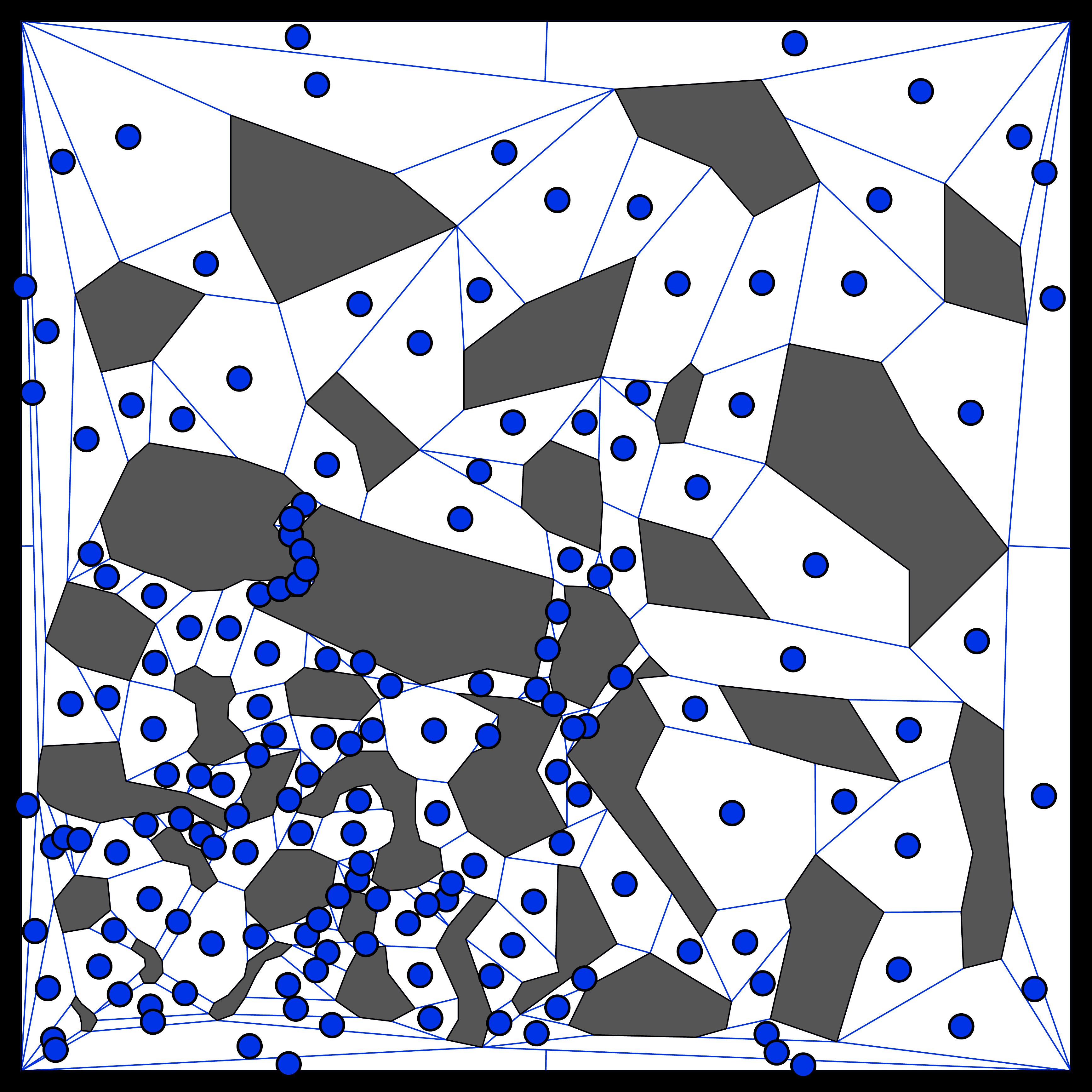}
        \caption*{KA}
        \caption*{(initial: 161)}
    \end{subfigure}
    \begin{subfigure}[t]{0.155\columnwidth}
        \centering
        \includegraphics[width=\linewidth]{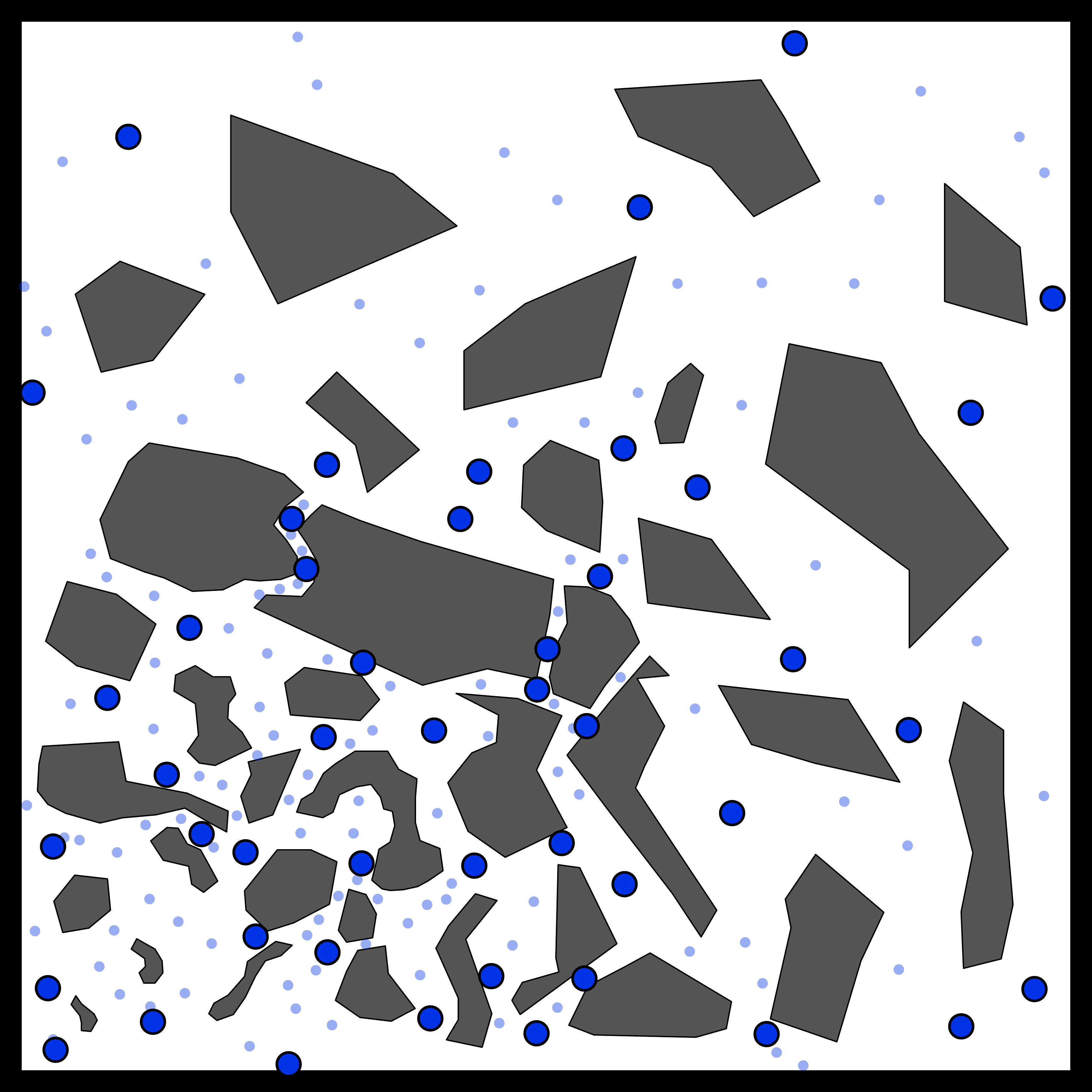}
        \caption*{HR-KA}
        \caption*{(refined: 46)}
    \end{subfigure}
    \begin{subfigure}[t]{0.155\columnwidth}
        \centering
        \includegraphics[width=\linewidth]{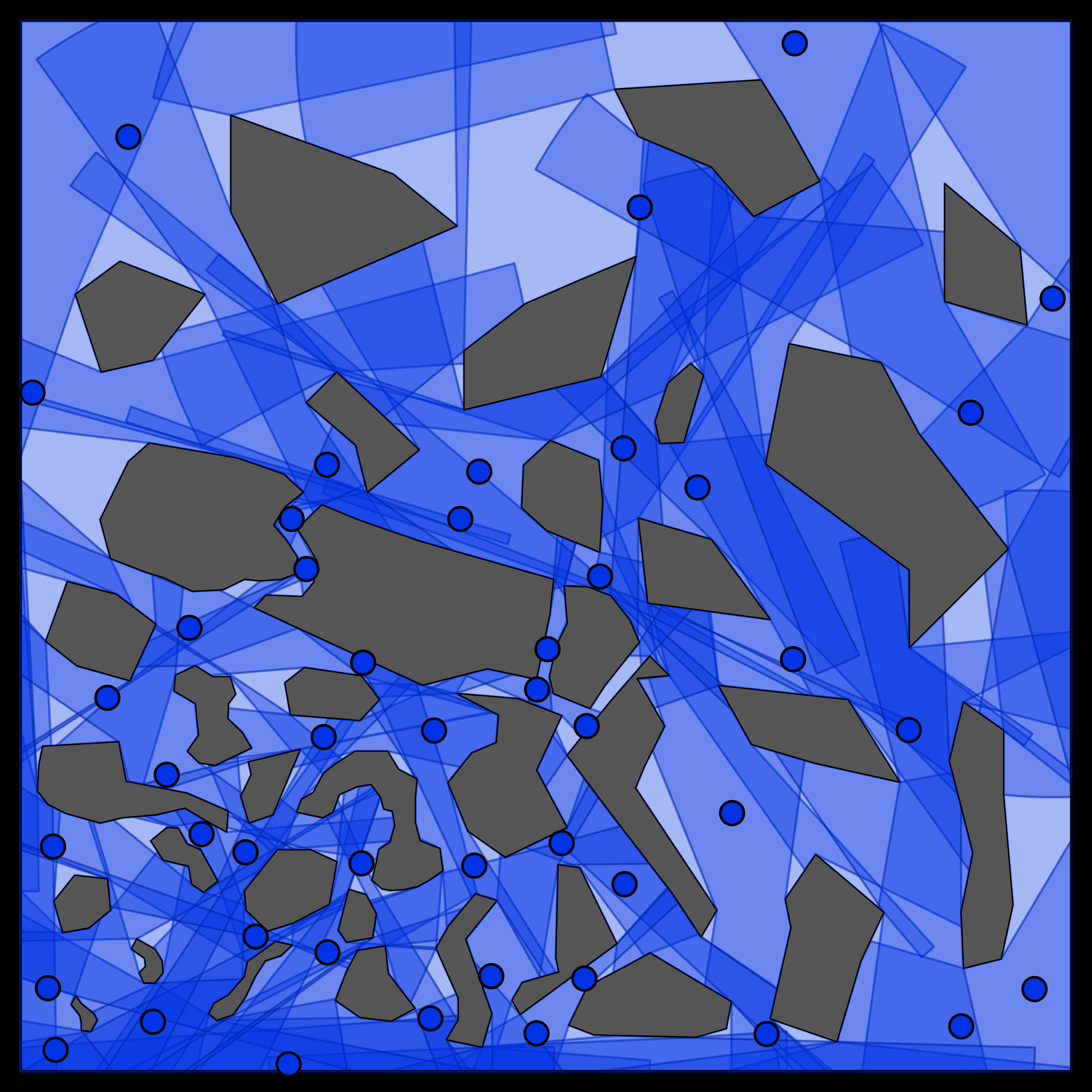}
        \caption*{HR-KA}
        \caption*{(coverage)}
    \end{subfigure}
    \begin{subfigure}[t]{0.155\columnwidth}
        \centering
        \includegraphics[width=\linewidth]{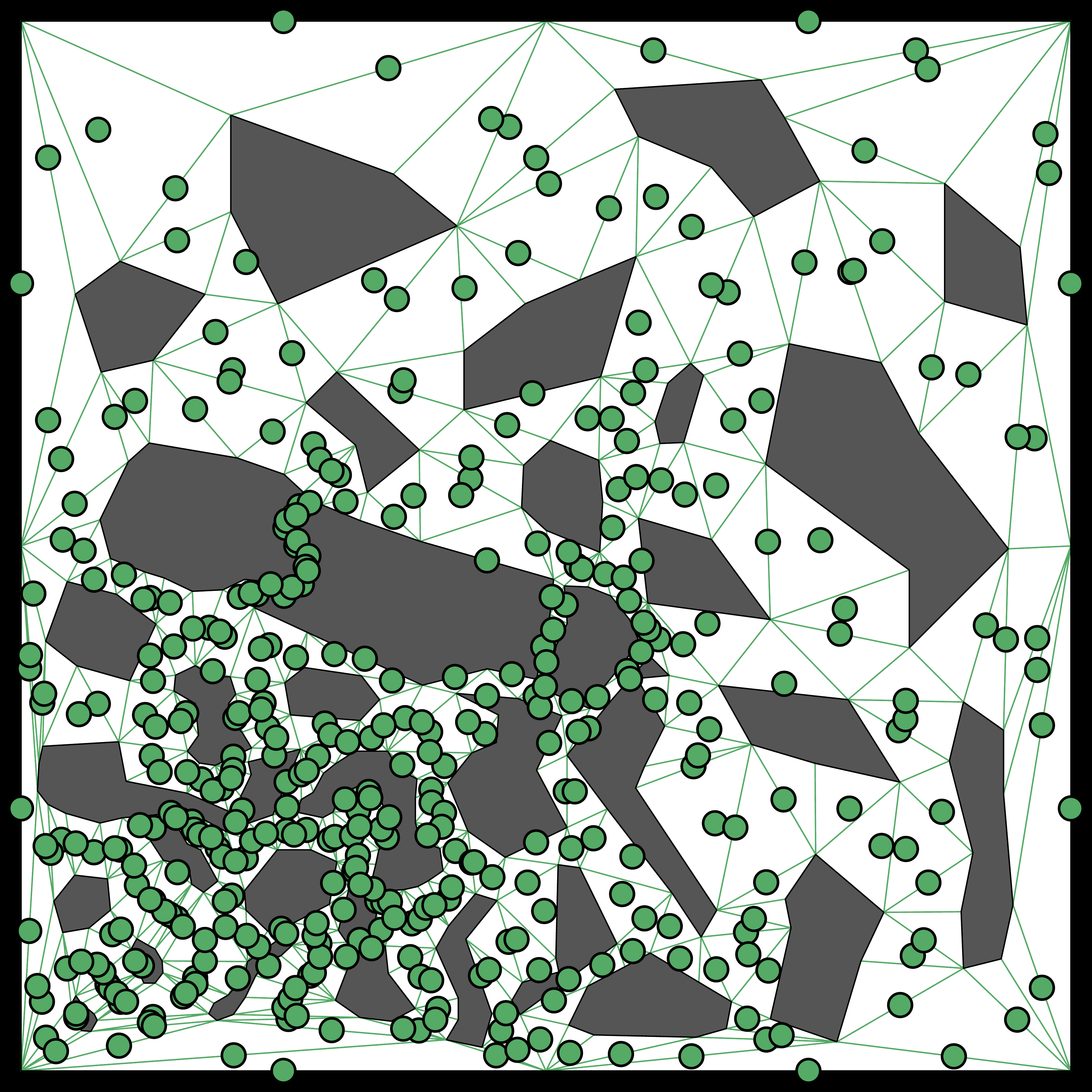}
        \caption*{CCDT}
        \caption*{(initial: 403)}
    \end{subfigure}
    \begin{subfigure}[t]{0.155\columnwidth}
        \centering
        \includegraphics[width=\linewidth]{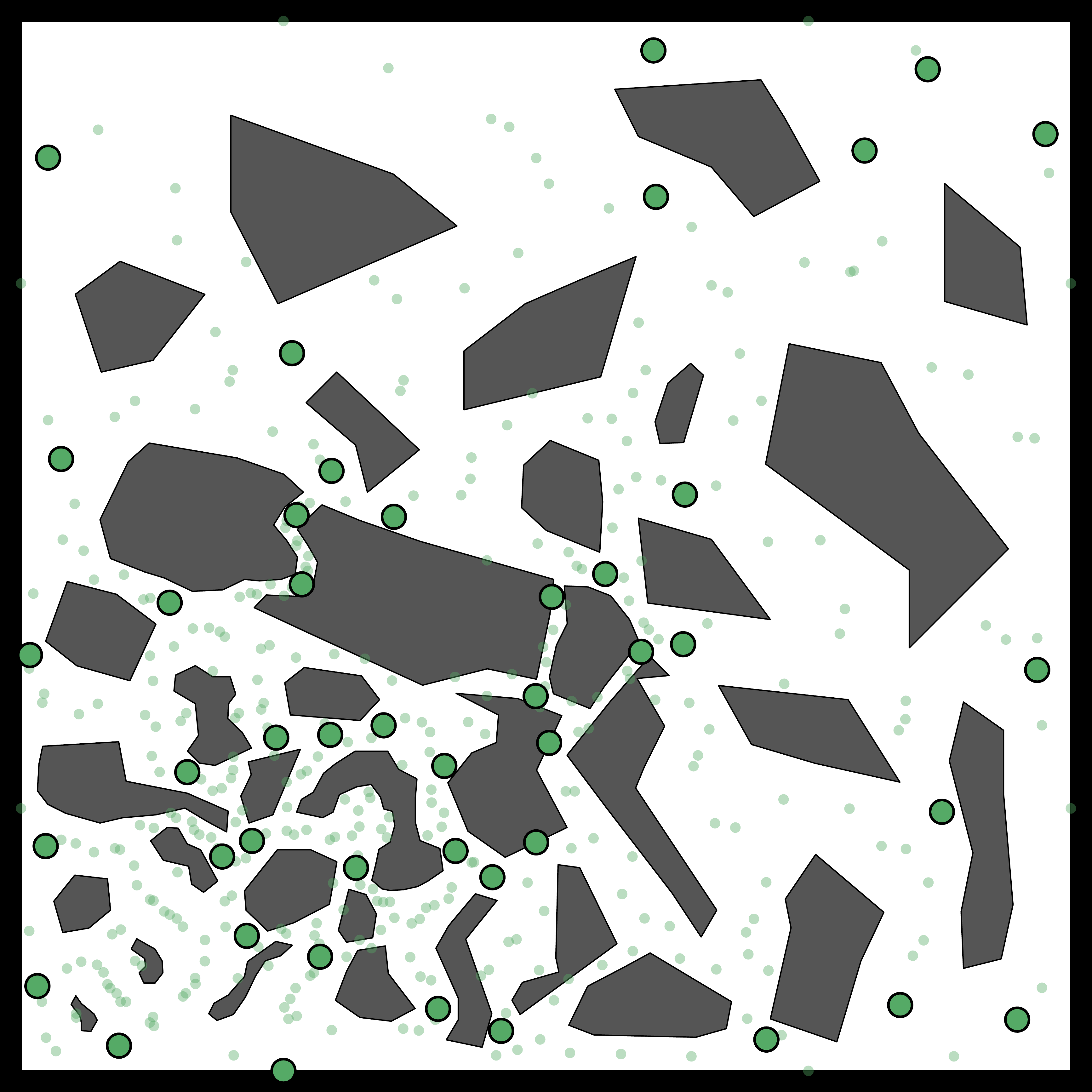}
        \caption*{HR-CCDT}
        \caption*{(refined: 45)}
    \end{subfigure}
    \begin{subfigure}[t]{0.155\columnwidth}
        \centering
        \includegraphics[width=\linewidth]{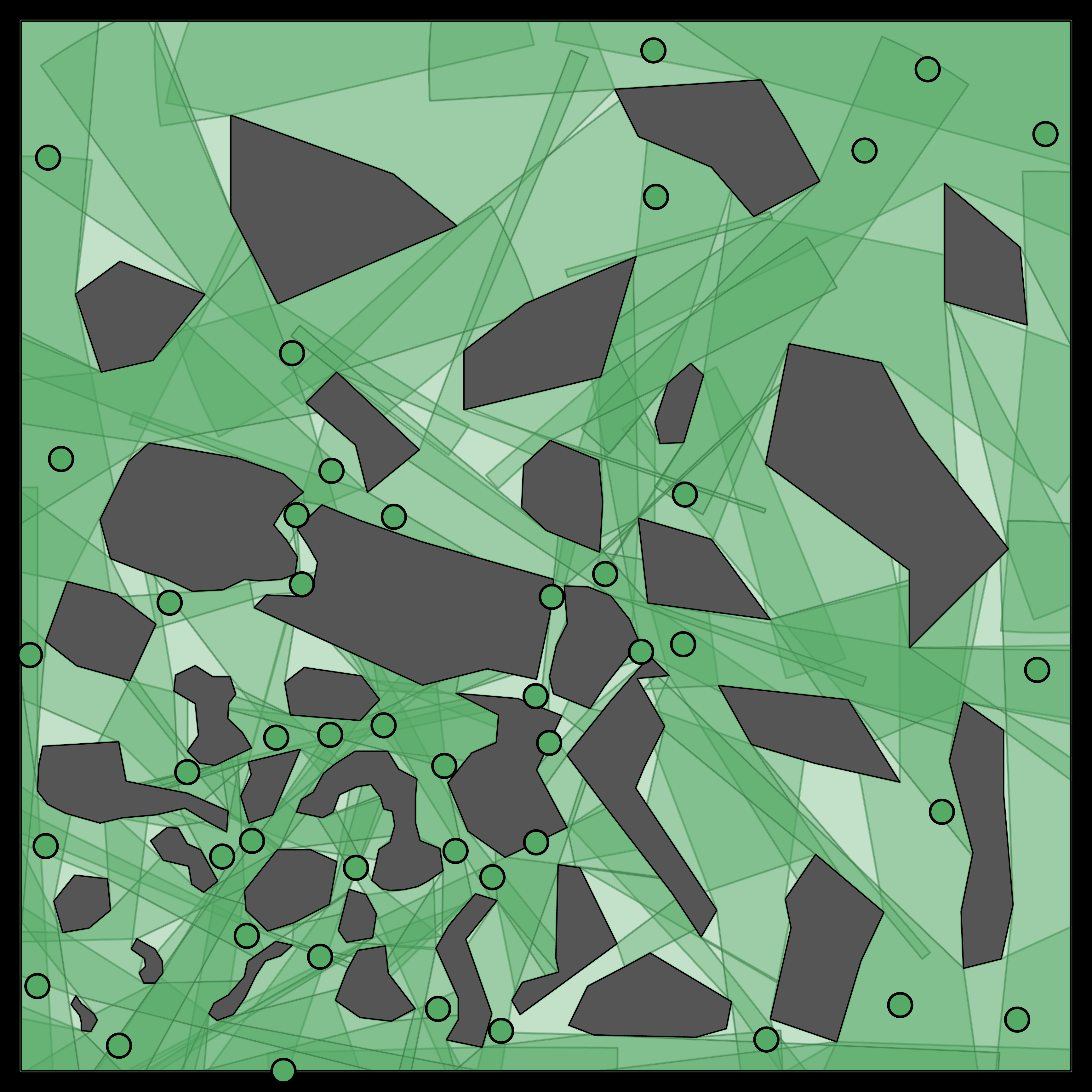}
        \caption*{HR-CCDT}
        \caption*{(coverage)}
    \end{subfigure}
    \\
    \begin{subfigure}[t]{0.155\columnwidth}
        \centering
        \includegraphics[width=\linewidth]{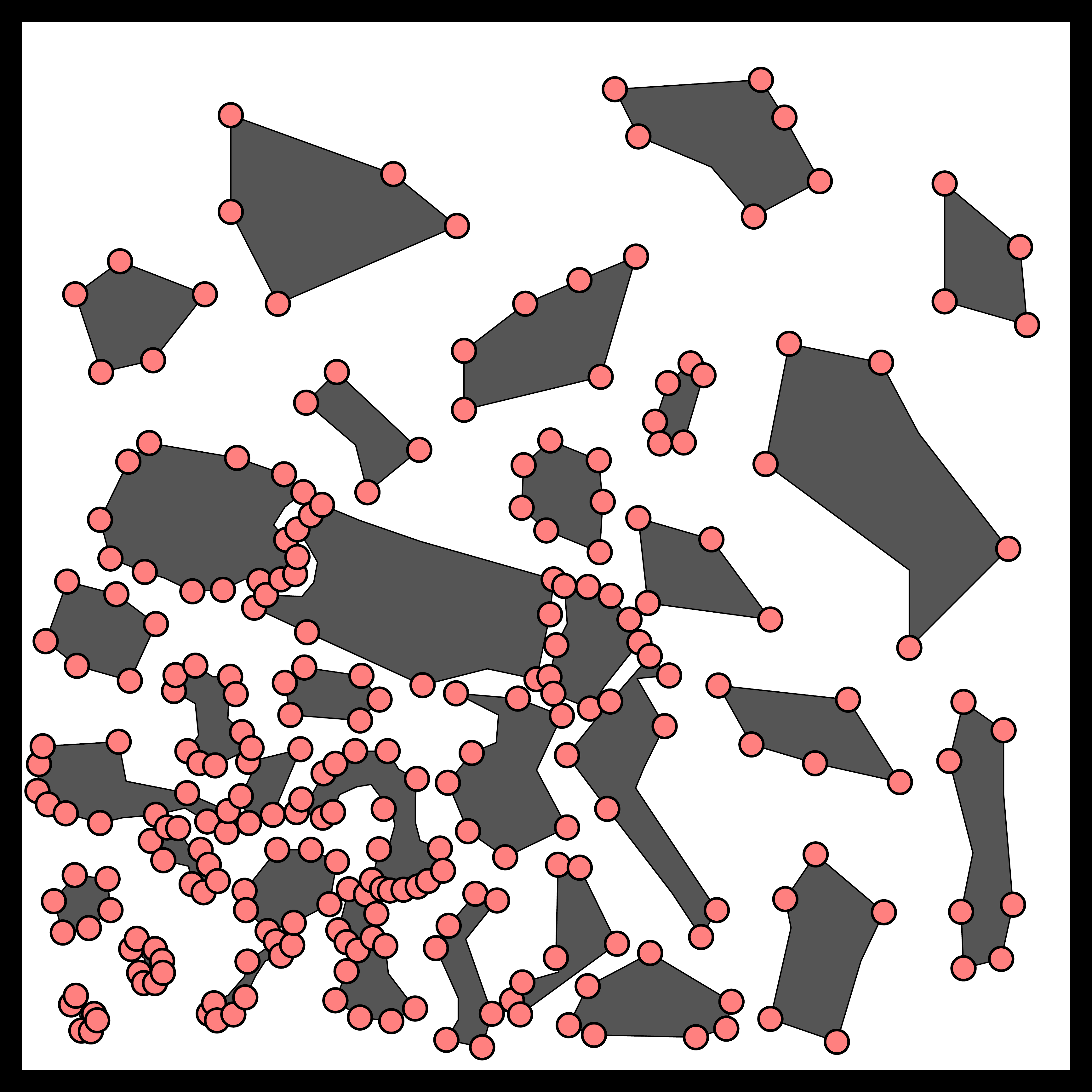}
        \caption*{RV}
        \caption*{(initial: 257)}
    \end{subfigure}
    \begin{subfigure}[t]{0.155\columnwidth}
        \centering
        \includegraphics[width=\linewidth]{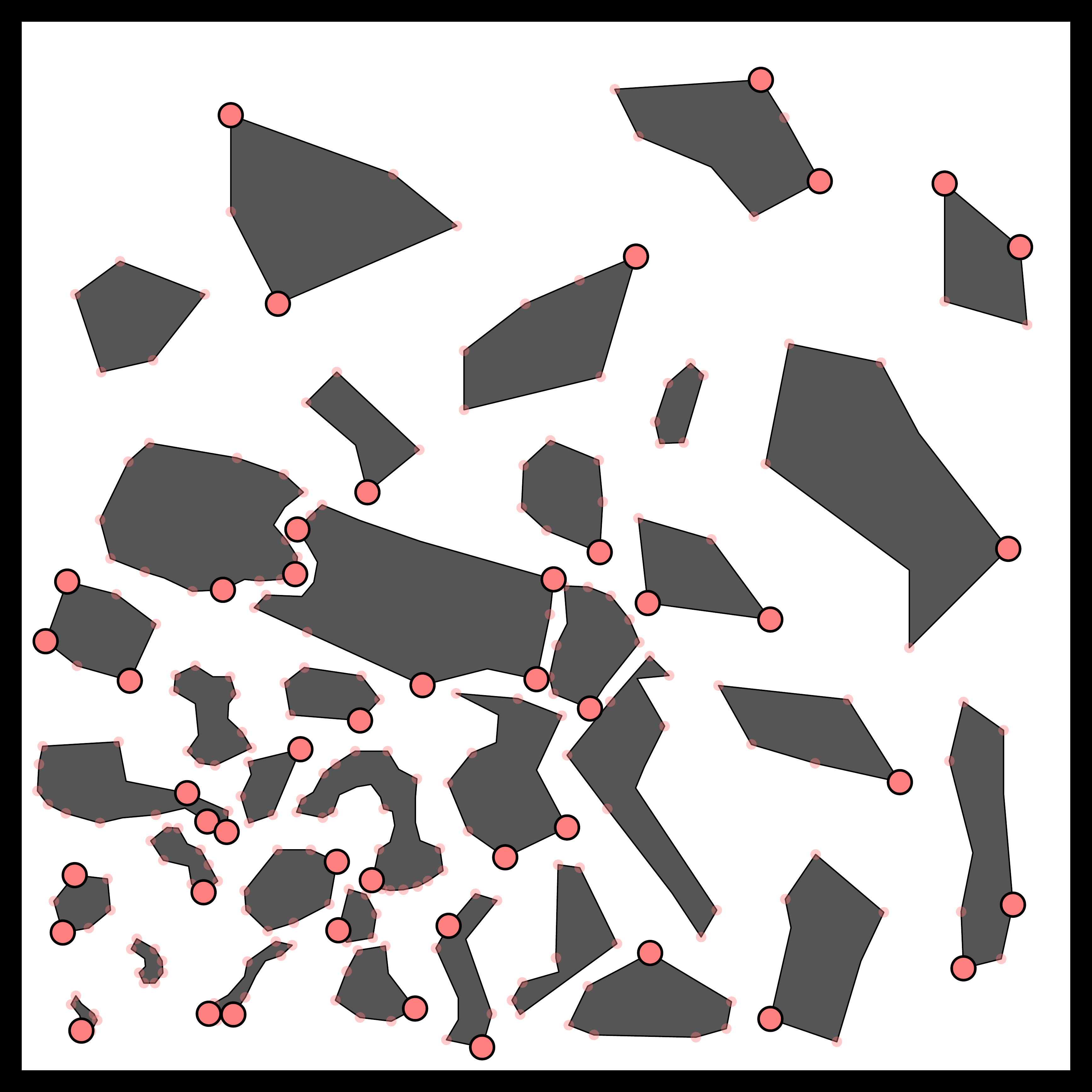}
        \caption*{HR-RV}
        \caption*{(refined: 46)}
    \end{subfigure}
    \begin{subfigure}[t]{0.155\columnwidth}
        \centering
        \includegraphics[width=\linewidth]{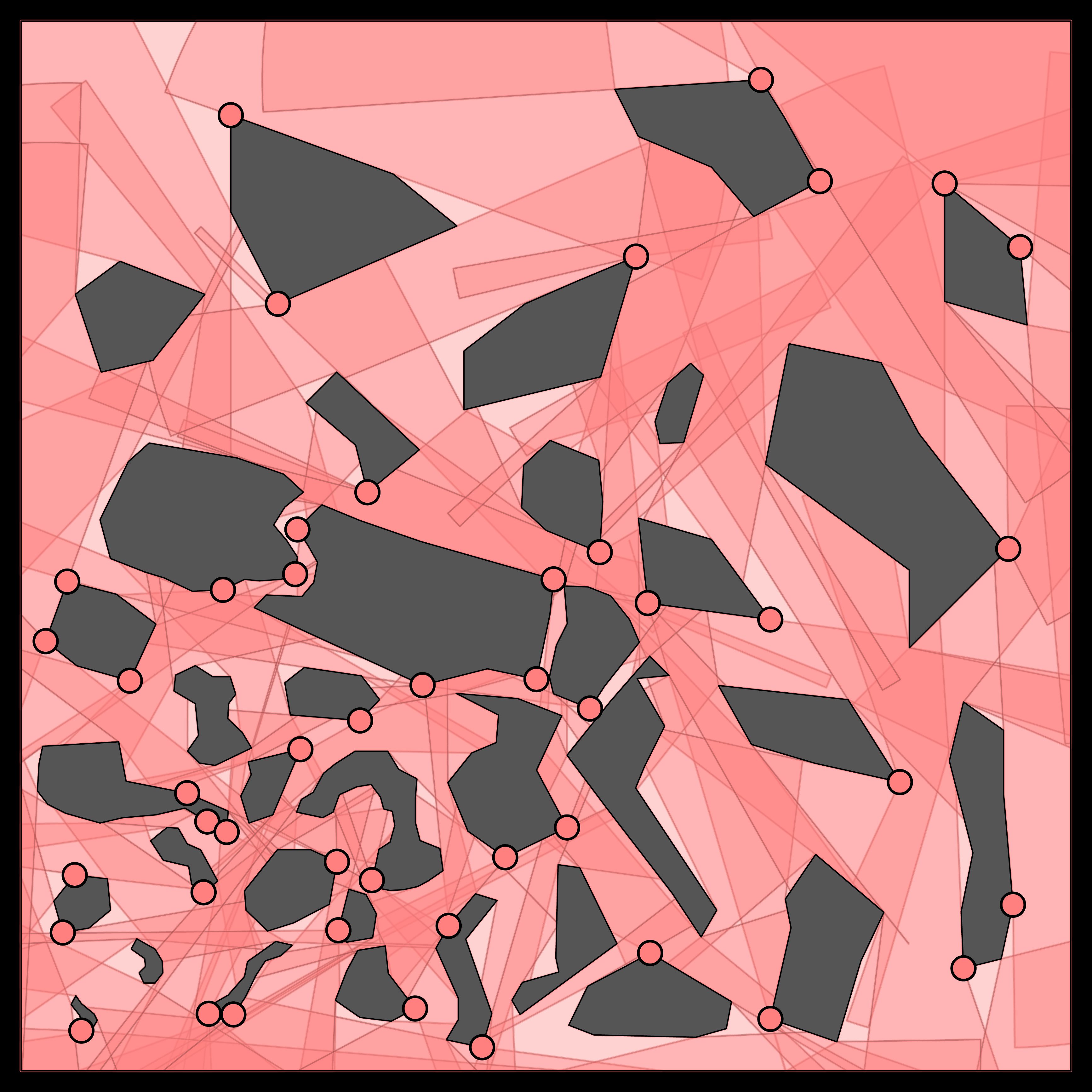}
        \caption*{HR-RV}
        \caption*{(coverage)}
    \end{subfigure}
    \begin{subfigure}[t]{0.155\columnwidth}
        \centering
        \includegraphics[width=\linewidth]{g/ka_ccdt_rv}
        \caption*{KA,CCDT,RV}
        \caption*{(combined: 821)}
    \end{subfigure}
    \begin{subfigure}[t]{0.155\columnwidth}
        \centering
        \includegraphics[width=\linewidth]{g/ka_ccdt_rv_filtered}
        \caption*{HR-KA,CCDT,RV}
        \caption*{(refined: 37)}
    \end{subfigure}
    \begin{subfigure}[t]{0.155\columnwidth}
        \centering
        \includegraphics[width=\linewidth]{g/ka_ccdt_rv_coverage}
        \caption*{HR-KA,CCDT,RV}
        \caption*{(coverage)}
    \end{subfigure}
    \caption{
        The \emph{hybrid refinement} (HR) framework follows two key principles: (1) combining outputs from multiple methods and (2) applying a refinement step.
        KA, CCDT, and RV are existing sensor-placement methods, while HR-KA,CCDT,RV represents their refined combination.
        Notably, HR-KA,CCDT,RV achieves the lowest guard count post-refinement.
        The final class of proposed methods, HAR heuristics, further improves efficiency by incorporating preprocessing techniques to accelerate refinement.
    }
    \label{fig:main-concepts}
\end{figure}

Our second contribution is the proposal of a new class of \emph{hybrid accelerated-refinement} (HAR) heuristics for the OSPP, which combine (hence \emph{hybrid}) and refine outputs from multiple existing sensor-placement methods while incorporating preprocessing techniques to accelerate the refinement step (hence \emph{accelerated-refinement}).
Despite being based on simple ideas, as illustrated in Fig.~\ref{fig:main-concepts}, an equivalent approach has not been explicitly proposed or evaluated for the SPP in the literature.

The remainder of this paper is organized as follows.
Sec.~\ref{sec:problem-statement} formally defines the problem, introduces the notation used throughout the paper, and presents relevant definitions, including the considered visibility models.
Sec.~\ref{sec:proposed-solution} introduces the new class of HAR heuristics, detailing the \emph{hybrid refinement}~(HR) framework and the accelerated-refinement step.
Sec.~\ref{sec:related-work} reviews related work and positions the proposed approach within the existing literature.
Sec.~\ref{sec:computational-evaluation} presents a comprehensive computational study of existing and proposed heuristics for the OSPP, evaluating them on a dataset of large, complex polygonal environments while analyzing the trade-off between sensor count and runtime efficiency.
Finally, Sec.~\ref{sec:conclusions} concludes the paper and outlines potential directions for future research.

\section{Problem Formulation and Related Definitions}
\label{sec:problem-statement}

\subsection{Notation for Sets and Spatial Representations}
\label{subsec:notation-sets}

Sets are denoted using normal-font and calligraphic uppercase Roman letters (e.g., $A$, $\mathcal{B}$).
The notation $|A|$ represents the cardinality of $A$.
We use $\{ . \cop{\mid} . \}$ for set-builder notation, $\emptyset$ for the empty set, $\setminus$ for set difference, $\cup$ for union, $\cap$ for intersection, and $\mathrm{cl}(.)$ for set closure.
The power set of $A$ is denoted by $2^A$, containing all subsets of $A$, including $\emptyset$ and $A$ itself.
Calligraphic letters (e.g., $\mathcal{A}, \mathcal{B}$) represent non-empty, bounded, closed, but not necessarily connected subsets of $\mathbb{R}^2$, while normal-font letters (e.g., $A, B$) denote finite sets of various objects (e.g., numbers, points, or other sets).
For spatial sets, $\partial\mathcal{A}$ denotes the boundary of $\mathcal{A}$, $\mathrm{Area}(\mathcal{A})$ gives its area, and $\mathrm{BoundBox}(\mathcal{A})$ is its smallest axis-aligned bounding box.
For finite sets, $\{a\}$ denotes a singleton, and $a_1 \ldots a_n \cop{\gets} A$ enumerates the elements of $A$, where $n \cop{=} |A|$.
The index $i$ of $a_i \cop{\in} A$ is referred to as its ID\@.

\subsection{Omnidirectional Sensor Placement Problem (General Formulation)}
\label{subsec:sensor-placement-problem}

The \emph{omnidirectional sensor placement problem} (OSPP) is a variant of the SPP that focuses on omnidirectional sensors with a general visibility model and a user-defined coverage ratio.
It assumes a \emph{connected environment} $\mathcal{W} \cop{\subset} \mathbb{R}^2$ and point sensors with configuration space $\mathbb{C} \cop{\coloneqq} \mathcal{W}$.
The \emph{visibility model} $\mathrm{Vis} \cop{:} \mathcal{W} \cop{\mapsto} 2^\mathcal{W}$ maps each point $p \cop{\in} \mathcal{W}$ to its \emph{visibility region} $\mathcal{V}_p \cop{\coloneqq} \mathrm{Vis}(p) \cop{\subset} \mathcal{W}$.
Obstacles are implicitly defined by $\mathcal{W}$, which represents the transparent space, while $\mathbb{R}^2 \cop{\setminus} \mathcal{W}$ represents occlusions.
The OSPP seeks the smallest finite set of sensor locations $G \cop{\subset} \mathcal{W}$ such that at least $(1 \cop{-} \epsilon)$ of $\mathcal{W}$ is visible:
\begin{equation}
    {\min}_{G \in 2^\mathcal{W}} |G| \quad \text{s.t.} \quad |G| \in \mathbb{N}, \quad \mathrm{Area}({\bigcup}_{g \in G}\mathrm{Vis}(g)) \geq (1 - \epsilon) \mathrm{Area}(\mathcal{W}),
    \label{eq:problem-definition}
\end{equation}
where $\epsilon \cop{\in} [0,1]$ allows for partial coverage to mitigate diminishing returns in complex environments.
Throughout this paper, $G$ is called the \emph{guard set}, with elements $g_1, \ldots, g_n$, where $n \cop{=} |G|$, referred to as \emph{guards}.
The set of guard-visibility pairs, $C \cop{\coloneqq} \{(g_1, \mathcal{V}_1), \ldots, (g_n, \mathcal{V}_n)\}$, where $\mathcal{V}_i \cop{\coloneqq} \mathrm{Vis}(g_i)$, defines the \emph{coverage} of $\mathcal{W}$ by $G$.

\paragraph*{Remark on the Problem Solvability}

The preceding formulation, particularly the requirement of a finite guard set $G$, assumes that the environment $\mathcal{W}$ and the visibility model $\mathrm{Vis}$ exhibit well-behaved properties.
For instance, if $\mathcal{W}$ had fractal boundaries or if $\mathrm{Vis}$ were restricted to a finite set of points, such as $\mathrm{Vis}(p) \cop{\coloneqq} \{ p \}$, the minimum guard set could be infinite, rendering the problem unsolvable.
To exclude such pathological cases, we assume throughout this paper that $\mathcal{W}$ and $\mathrm{Vis}$ are sufficiently well-behaved to guarantee a finite solution.
A rigorous theoretical discussion of solvability conditions is beyond the scope of this practically oriented work.

\subsection{Polygonal Representation of the Environment}
\label{subsec:polygonal-domains}

The general problem formulation, along with the proposed solution framework introduced later, applies to general 2D environments with arbitrary boundary shapes, including polygonal, rectilinear, and smooth curves.
However, our implementation and evaluation focus specifically on polygonal environments.
A \emph{polygonal environment} $\mathcal{W}$ is defined by boundaries composed of line segments forming \emph{simple polygons}.
A \emph{polygon} is a \emph{closed polygonal chain}, and \emph{simple} refers to the property of being strictly \emph{non-self-intersecting}.
The environment $\mathcal{W}$ has a single \emph{outer boundary} and zero or more \emph{inner boundaries}, called \emph{holes}, with all boundaries being pairwise \emph{weakly non-intersecting}.
\emph{Weakly non-intersecting} means that while boundaries may touch at isolated points or segments, they cannot cross each other.
If two boundaries share a segment, they can be merged into a single boundary.

\subsection{Visibility Model Definitions}
\label{subsec:visibility-models}

\begin{figure}[t]
    \centering
    \begin{subfigure}[t]{0.155\columnwidth}
        \centering
        \includegraphics[width=\linewidth]{g/vis_inf}
        \caption*{$\mathrm{Vis}_{\infty}$}
        \label{fig:vis-inf}
    \end{subfigure}
    \begin{subfigure}[t]{0.155\columnwidth}
        \centering
        \includegraphics[width=\linewidth]{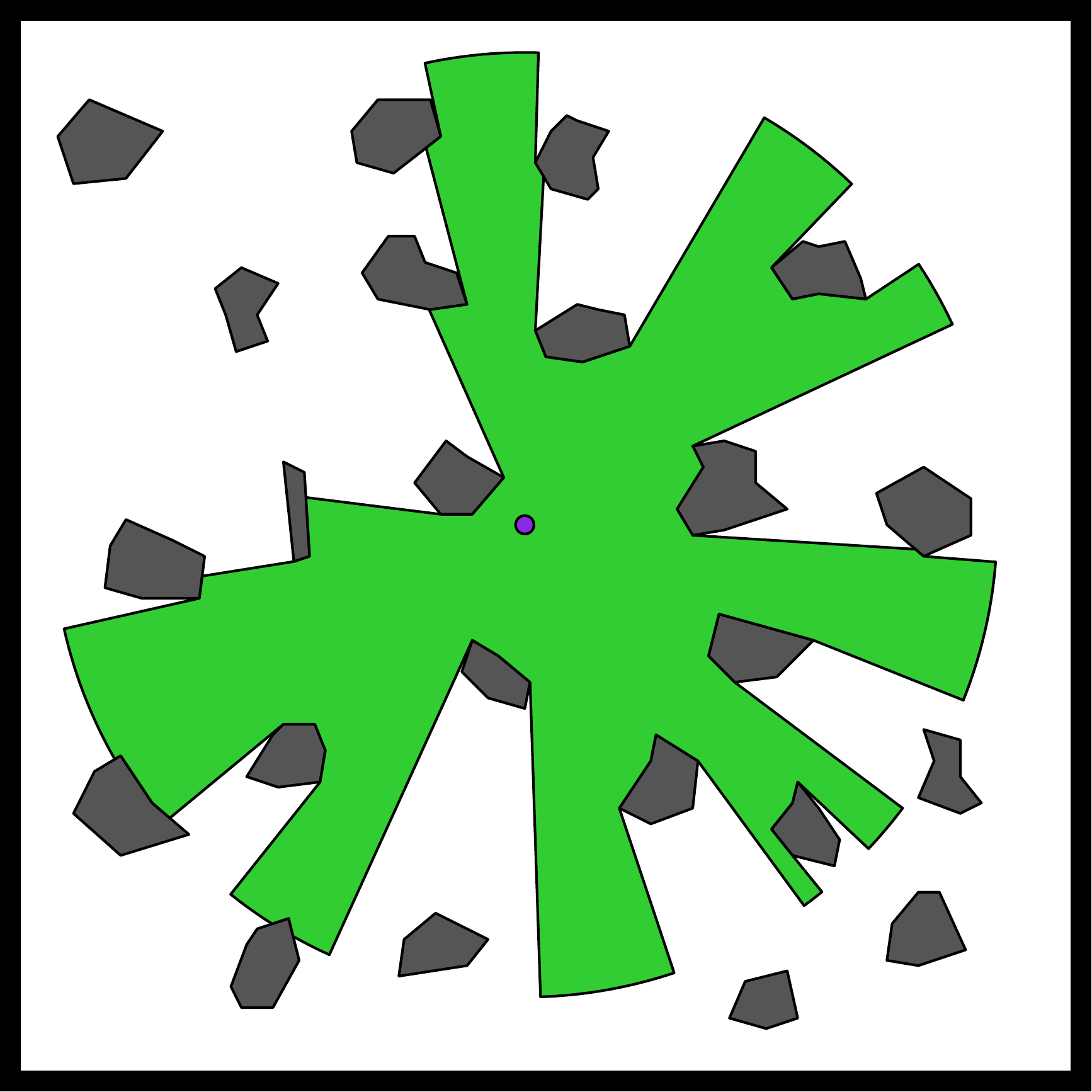}
        \caption*{$\mathrm{Vis}_{d=9}$}
    \end{subfigure}
    \begin{subfigure}[t]{0.155\columnwidth}
        \centering
        \includegraphics[width=\linewidth]{g/vis_45}
        \caption*{$\mathrm{Vis}_{d=4.5}$}
    \end{subfigure}
    \begin{subfigure}[t]{0.155\columnwidth}
        \centering
        \includegraphics[width=\linewidth]{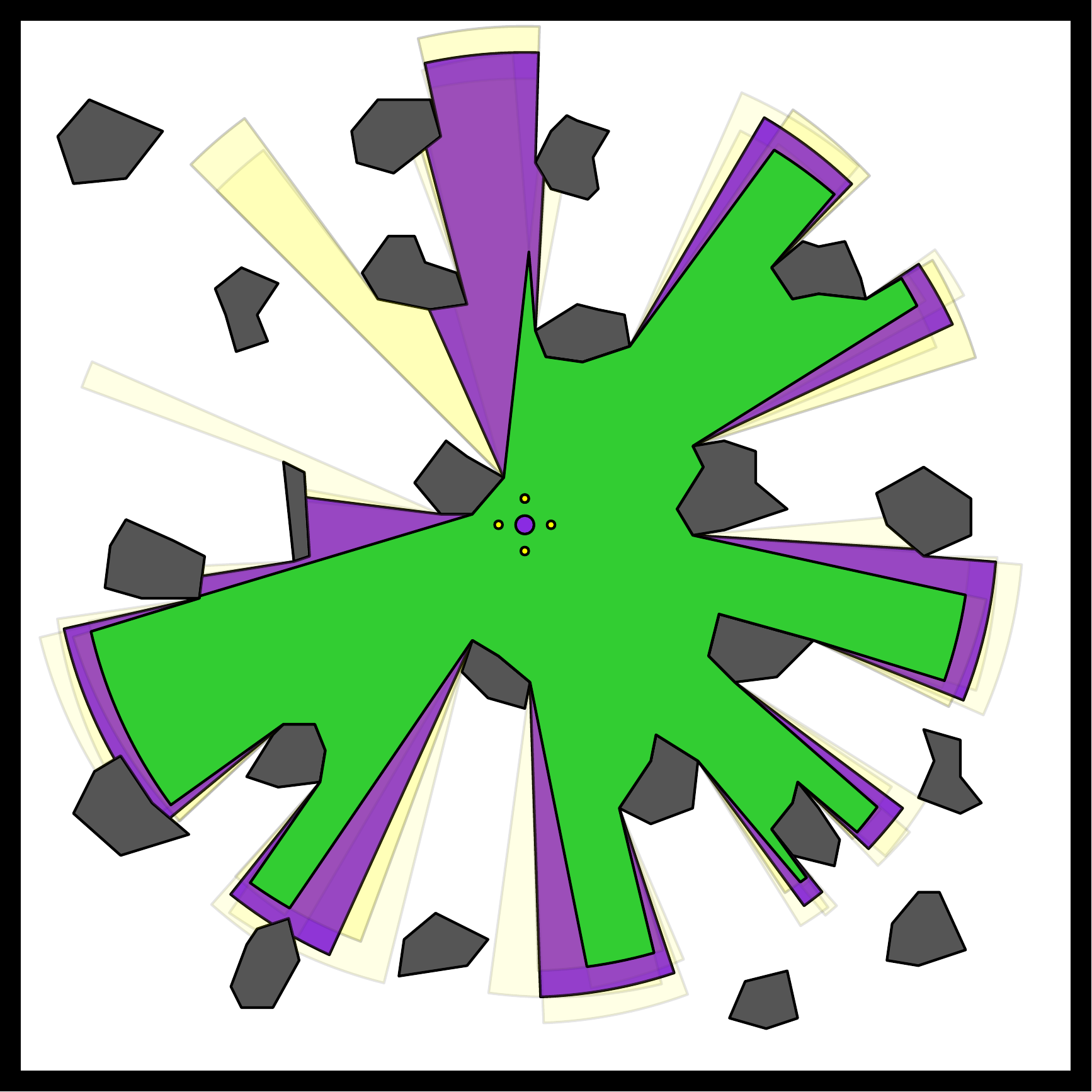}
        \caption*{$\mathrm{Vis}_{d=9}, \mathrm{Unc}_{r=0.5}$}
        \caption*{$r_{\text{samp}} \cop{\approx} 0.785$}
    \end{subfigure}
    \begin{subfigure}[t]{0.155\columnwidth}
        \centering
        \includegraphics[width=\linewidth]{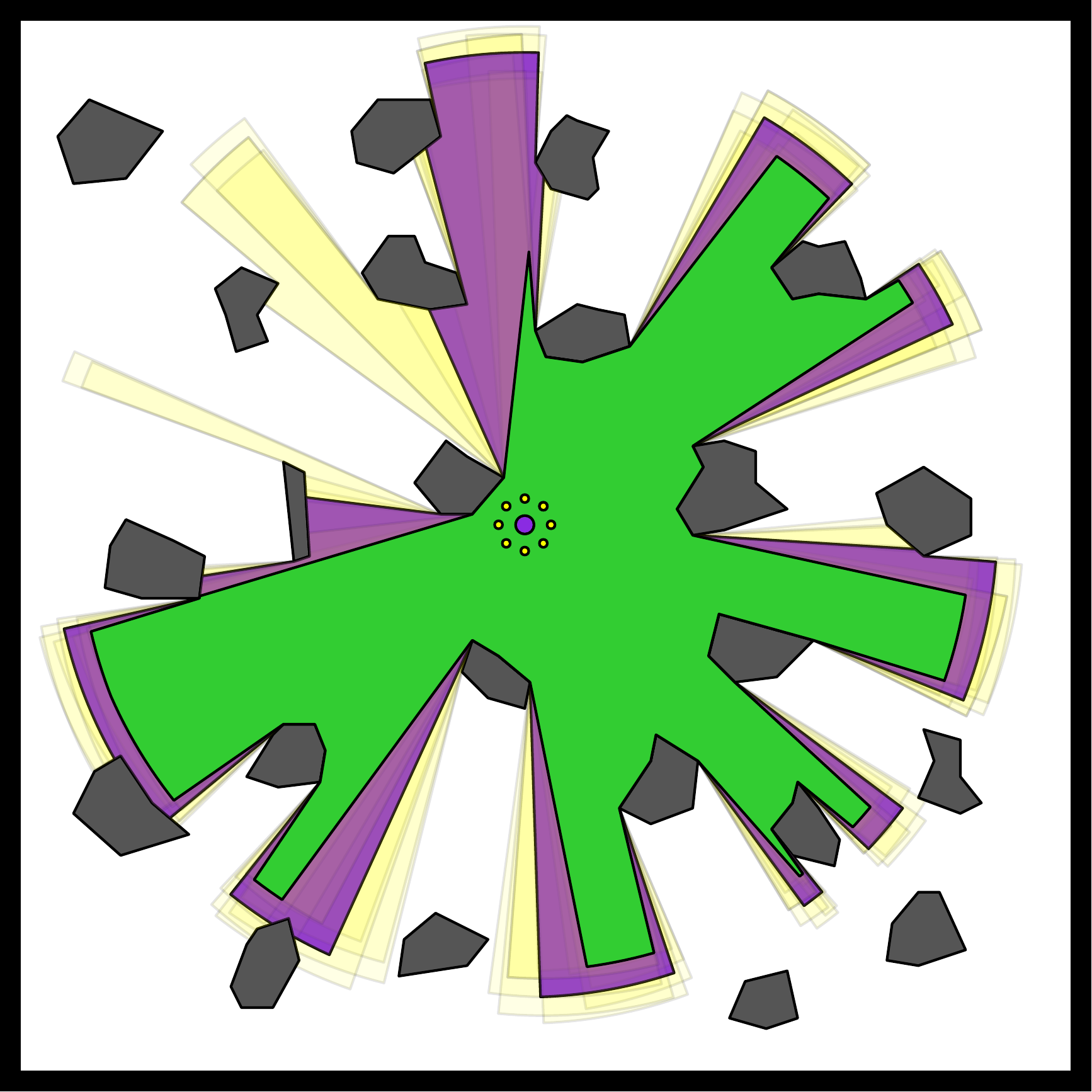}
        \caption*{$\mathrm{Vis}_{d=9}, \mathrm{Unc}_{r=0.5}$}
        \caption*{$r_{\text{samp}} \cop{\approx} 0.393$}
    \end{subfigure}
    \begin{subfigure}[t]{0.155\columnwidth}
        \centering
        \includegraphics[width=\linewidth]{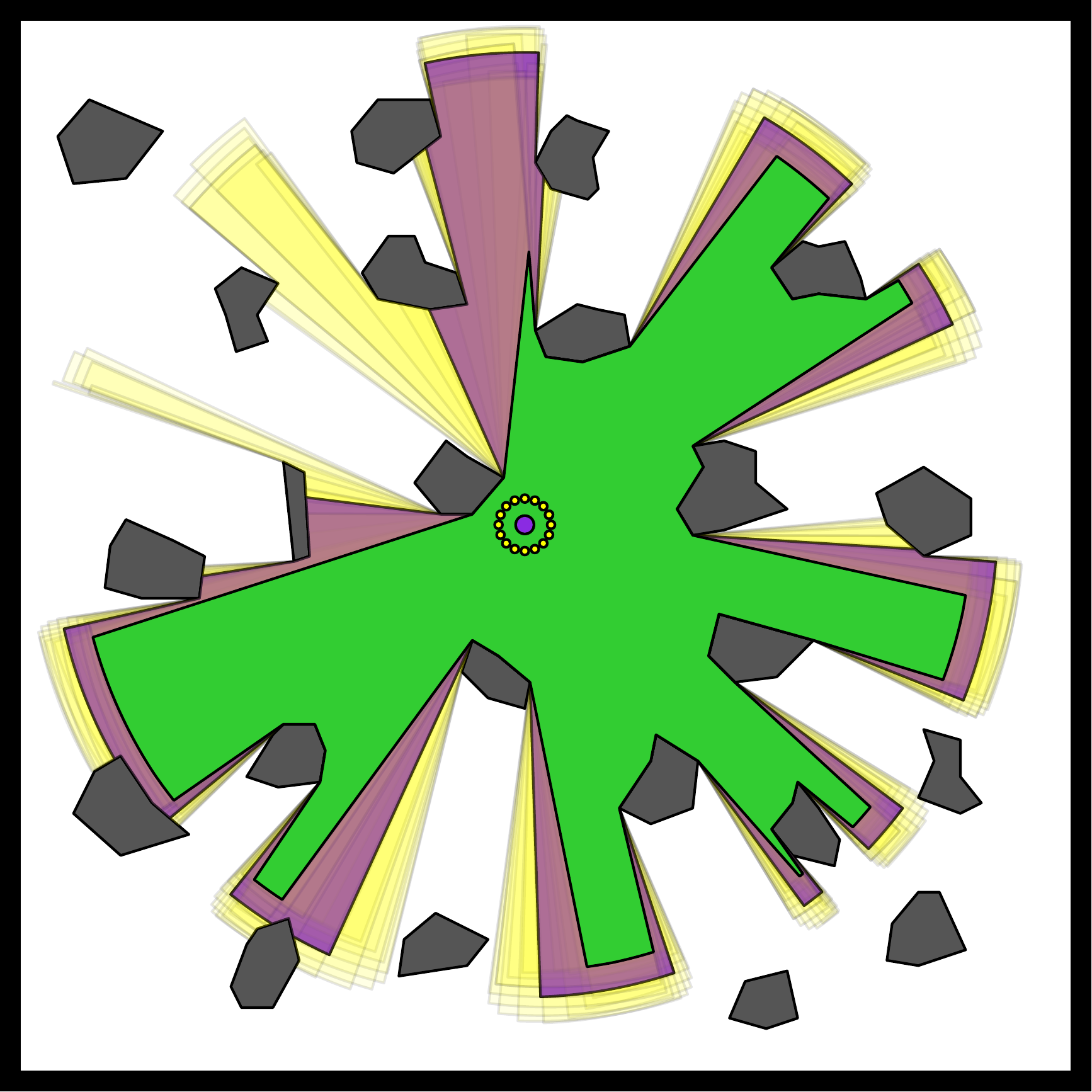}
        \caption*{$\mathrm{Vis}_{d=9}, \mathrm{Unc}_{r=0.5}$}
        \caption*{$r_{\text{samp}} \cop{\approx} 0.196$}
    \end{subfigure}
    \\
    \begin{subfigure}[t]{0.155\columnwidth}
        \centering
        \includegraphics[width=\linewidth]{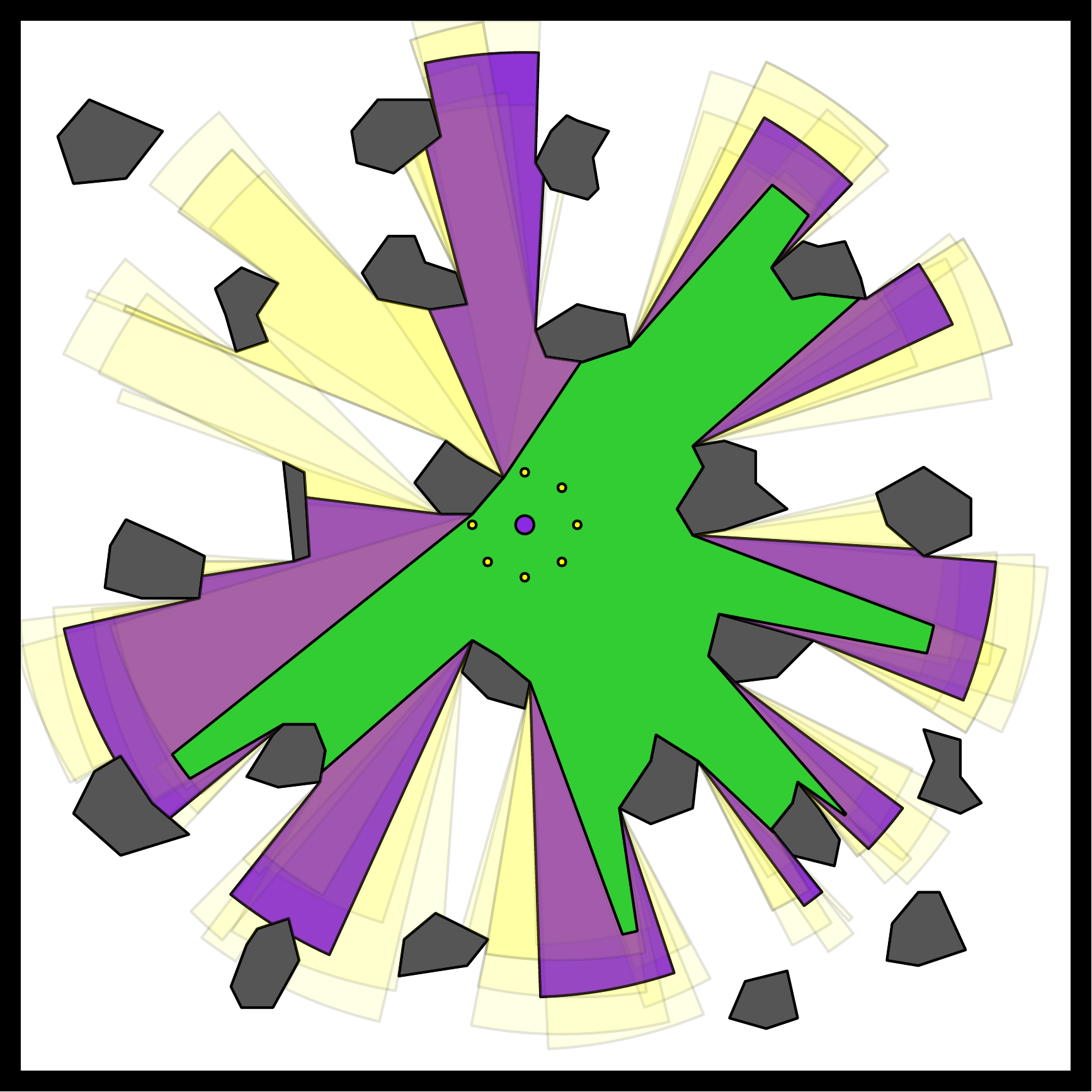}
        \caption*{$\mathrm{Vis}_{d=9}, \mathrm{Unc}_{r=1}$}
        \caption*{$r_{\text{samp}} \cop{\approx} 0.785$}
    \end{subfigure}
    \begin{subfigure}[t]{0.155\columnwidth}
        \centering
        \includegraphics[width=\linewidth]{g/ex_vis_region_robust_potholes_9.600000-10.400000_r-1.000000_d-9.000000_sa-0.021817_ra-0.392699}
        \caption*{$\mathrm{Vis}_{d=9}, \mathrm{Unc}_{r=1}$}
        \caption*{$r_{\text{samp}} \cop{\approx} 0.393$}
    \end{subfigure}
    \begin{subfigure}[t]{0.155\columnwidth}
        \centering
        \includegraphics[width=\linewidth]{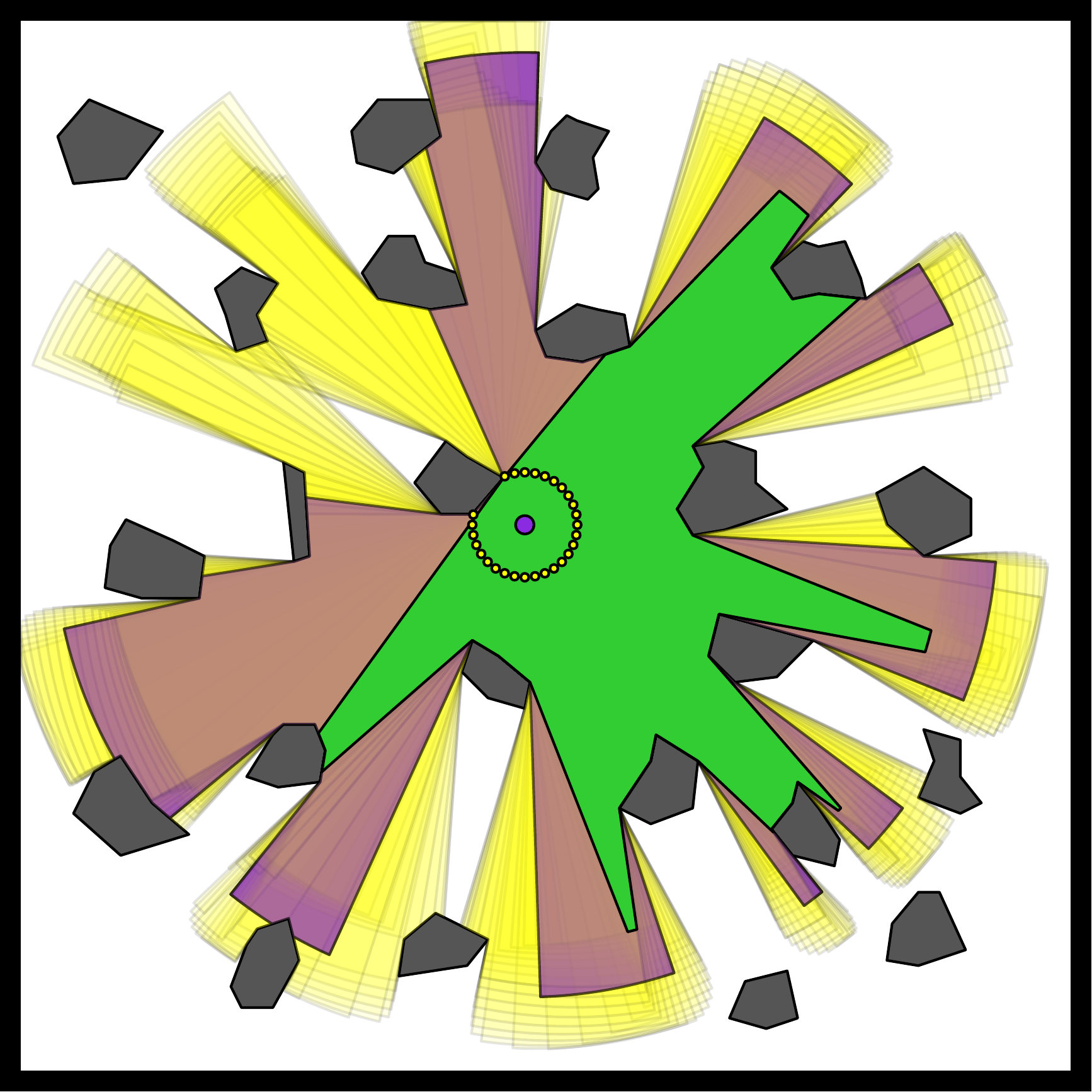}
        \caption*{$\mathrm{Vis}_{d=9}, \mathrm{Unc}_{r=1}$}
        \caption*{$r_{\text{samp}} \cop{\approx} 0.196$}
    \end{subfigure}
    \begin{subfigure}[t]{0.155\columnwidth}
        \centering
        \includegraphics[width=\linewidth]{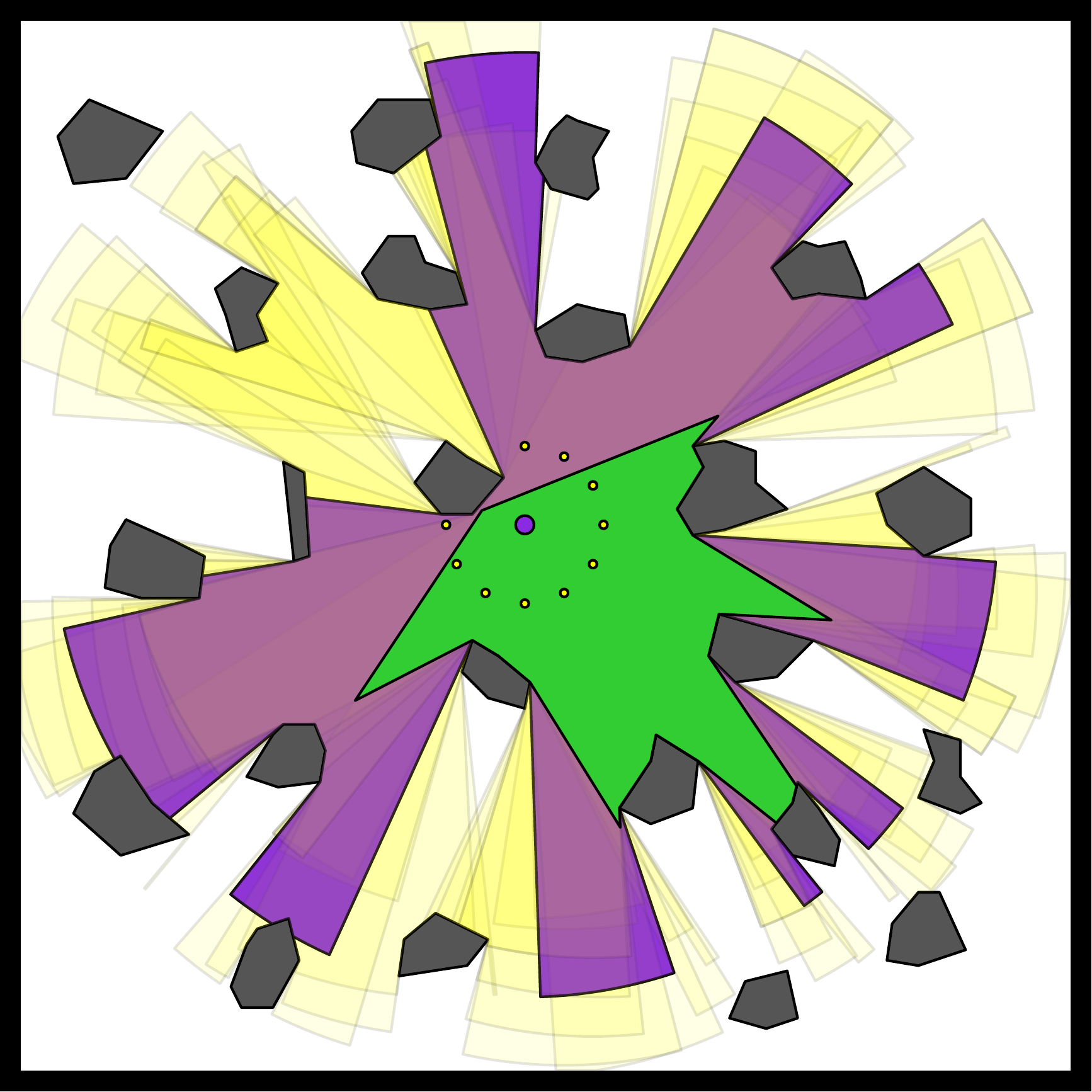}
        \caption*{$\mathrm{Vis}_{d=9}, \mathrm{Unc}_{r=1.5}$}
        \caption*{$r_{\text{samp}} \cop{\approx} 0.785$}
    \end{subfigure}
    \begin{subfigure}[t]{0.155\columnwidth}
        \centering
        \includegraphics[width=\linewidth]{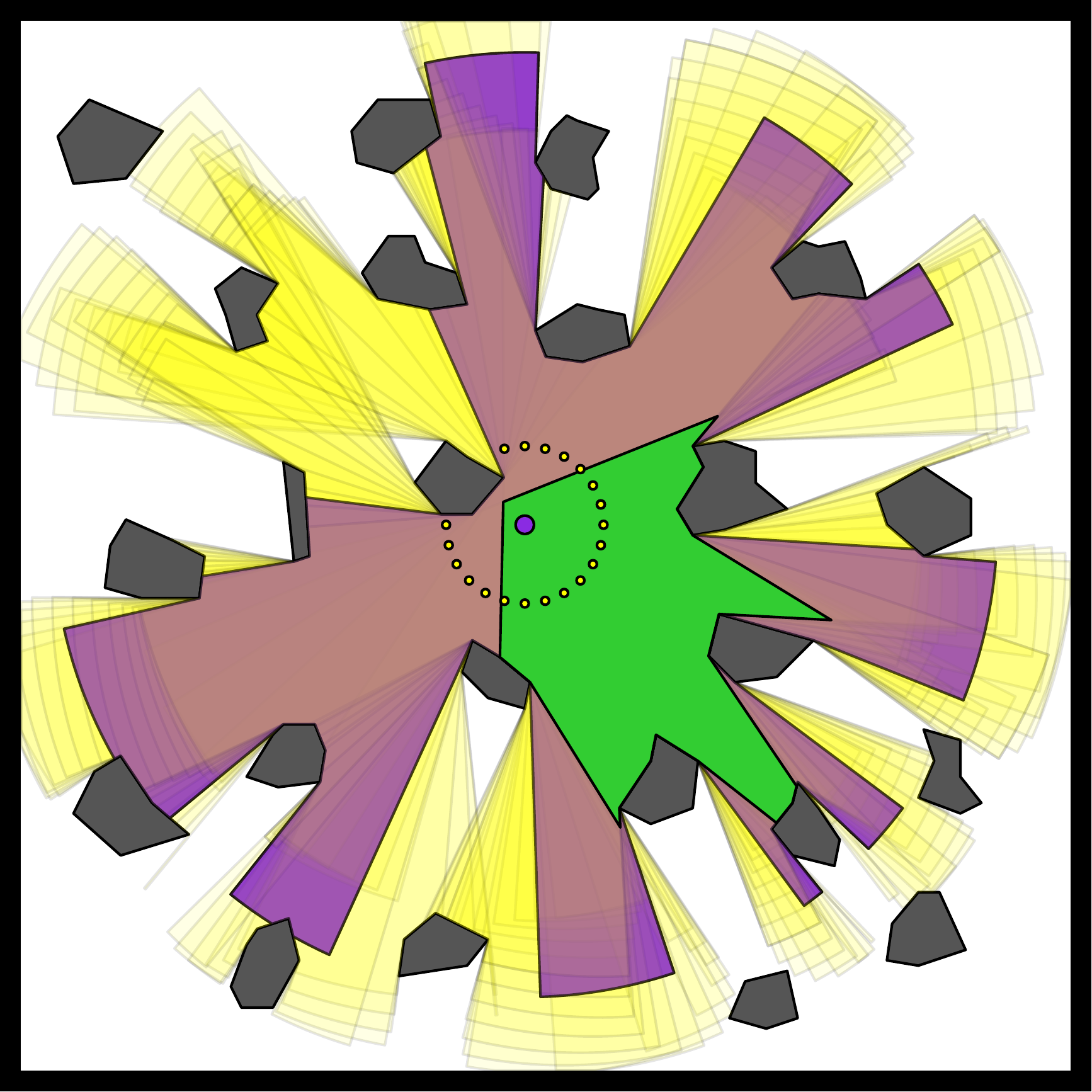}
        \caption*{$\mathrm{Vis}_{d=9}, \mathrm{Unc}_{r=1.5}$}
        \caption*{$r_{\text{samp}} \cop{\approx} 0.393$}
    \end{subfigure}
    \begin{subfigure}[t]{0.155\columnwidth}
        \centering
        \includegraphics[width=\linewidth]{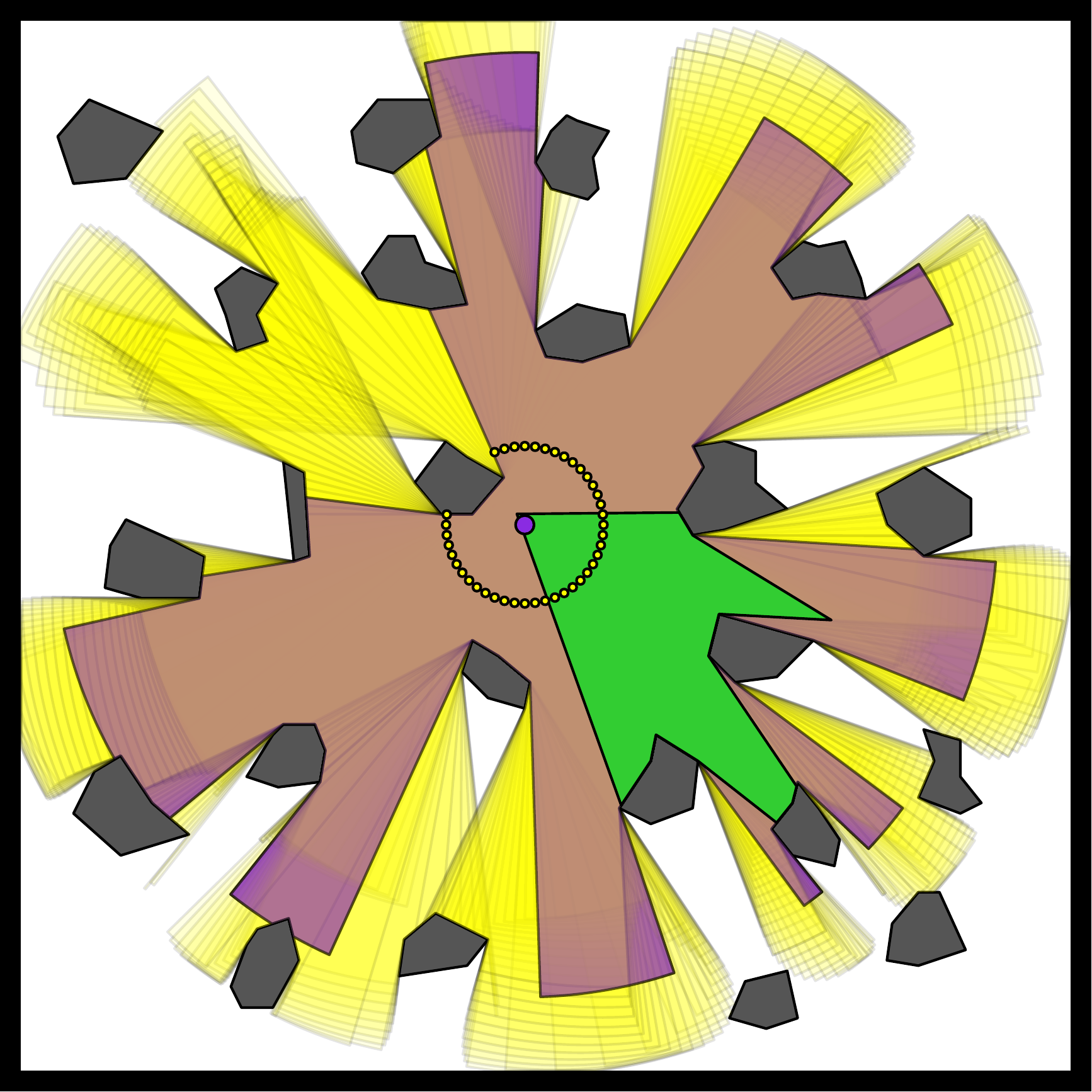}
        \caption*{$\mathrm{Vis}_{d=9}, \mathrm{Unc}_{r=1.5}$}
        \caption*{$r_{\text{samp}} \cop{\approx} 0.196$}
    \end{subfigure}
    \caption{
        Examples of visibility models for the same guard location in a $20\,\mathrm{m} \cop{\times} 20\,\mathrm{m}$ polygonal environment.
        Visibility regions are shown in green, with the guard represented as a central violet dot.
        For the localization-uncertainty models, the guard's own visibility region is displayed in semi-transparent violet, while the sampled points on the uncertainty region boundary form rings around the guard, with their visibility regions shown in semi-transparent yellow.
        All parameters are in meters.
    }
    \label{fig:visibility-models}
\end{figure}

Our computational study examines three visibility models.
The following paragraphs detail each model and conclude with practical considerations related to \emph{region clipping} operations.
Fig.~\ref{fig:visibility-models} provides a visual overview of the discussed models.

\paragraph{Unlimited Visibility}

The \emph{unlimited omnidirectional visibility} model in a polygonal domain is defined as:
\begin{equation}
    \mathrm{Vis}_{\infty}(g) \coloneqq \{ p \in \mathcal{W} \mid \overline{gp} \subset \mathcal{W} \},
    \label{eq:visibility-model}
\end{equation}
where $\overline{gp}$ denotes the line segment connecting $g$ and $p$.
The visibility region around $g$ forms a \emph{star-shaped polygon}, potentially with one-dimensional \emph{antennas}~\citep{Bungiu2014}.
An antenna arises when $g$ aligns with two visible vertices of the environment, restricting visibility from opposite sides.
Since antennas have zero area, they do not affect coverage quality and can be ignored.
Thus, visibility regions defined by Eq.~\eqref{eq:visibility-model} are effectively \emph{simple polygons}.
For a polygonal environment without holes, with $\epsilon \cop{=} 0$, and the visibility model in Eq.~\eqref{eq:visibility-model}, the OSPP reduces to the NP-hard \emph{art gallery problem} (AGP)~\citep{Chvatal1975}.
By this reduction, OSPP is at least as hard as AGP\@.

\paragraph{Limited-Range Visibility}

To model real-world sensor limitations, we consider the \emph{limited-range visibility} model, where a guard $g$ can only see within a fixed distance $d \cop{\in} \mathbb{R}^+$:
\begin{equation}
    \mathrm{Vis}_{d}(g) \coloneqq \{ p \in \mathcal{W} \mid \overline{gp} \subset \mathcal{W} \land \|g - p\| \leq d \},
    \label{eq:limited-visibility-model}
\end{equation}
where $\|.\|$ denotes the Euclidean norm.

\paragraph{Localization-Uncertainty Visibility}

To model scenarios where a sensor's position is uncertain, we introduce the \emph{localization-uncertainty visibility} model.
Beyond localization errors, this uncertainty may arise from actuator imprecision or control inaccuracies.
Instead of a single point, the sensor's position is represented as a \emph{region}, defined by a \emph{localization-uncertainty model} $\mathrm{Unc} \cop{:} \mathcal{W} \cop{\mapsto} 2^\mathcal{W}$, which maps a point $p \cop{\in} \mathcal{W}$ to its \emph{uncertainty region} $\mathcal{U} \cop{\coloneqq} \mathrm{Unc}(p) \cop{\subset} \mathcal{W}$.
The \emph{localization-uncertainty visibility} model extends a given visibility model $\mathrm{Vis}$ by defining a sensor's visibility as the intersection of all visibility regions corresponding to its possible locations:
\begin{equation}
    \mathrm{Vis}_{\mathrm{Unc}}(g) \coloneqq {\bigcap}_{u \in \mathrm{Unc}(g)} \mathrm{Vis}(u).
    \label{eq:visibility-model-uncertainty}
\end{equation}
In our computational study, we adopt a simple uncertainty model with a constant uncertainty level, equivalent to the $r$-limited visibility model, i.e., $\mathrm{Unc}_r \cop{\coloneqq} \mathrm{Vis}_r$.
This represents localization error as a maximum distance $r$ between the sensor's actual position $p$ and the guard's location $g$, with the additional constraint that $p$ and $g$ must be mutually visible.
This constraint is crucial, as without it, the uncertainty region could be fragmented by nearby obstacles into multiple disconnected parts that are not visible to each other, resulting in a null visibility region under Eq.~\eqref{eq:visibility-model-uncertainty}, making the guard completely blind.
We define the combined model $\mathrm{Vis}_{d,\mathrm{Unc}_r}$ by integrating $\mathrm{Unc}_r$ with the $d$-limited visibility model, $\mathrm{Vis}_d$, where $d \cop{\in} (r, \infty]$.
In obstacle-free conditions within distance $r$ of $g$, the uncertainty region forms a circle of radius $r$.
If the nearest obstacle is at least $d$ away, the visibility region is also a circle centered at $g$, but with radius $d \cop{-} r$.
However, when obstacles are nearby, the situation becomes more complex, as illustrated in Fig.~\ref{fig:visibility-models}.

\paragraph{Practical Considerations Regarding Clipping Operations}

Our OSPP implementation for polygonal environments inherently requires \emph{region-clipping} operations (see the union in Eq.~\eqref{eq:problem-definition} and the intersection in Eq.~\eqref{eq:visibility-model-uncertainty}).
For the unlimited visibility model, where visibility regions are simple polygons, this process is straightforward due to existing polygon-clipping libraries~\citep{Clipper2}.
In contrast, the limited-range visibility model introduces circular arcs, complicating clipping operations.
To address this, we approximate visibility regions as polygons by sampling each circular arc at equidistant points along its circumference, using a parameter $d_{\text{samp}}$ to ensure a maximum spacing of $d_{\text{samp}}' \cop{\leq} d_{\text{samp}}$.
The localization-uncertainty visibility model is even more challenging, as it requires intersecting infinitely many visibility regions—an intractable task with standard algorithms.
To approximate this, we consider a finite set of sample points, including the guard itself and equidistant samples along the uncertainty region’s boundary.
These samples are spaced at most $r_{\text{samp}}$ apart, with those not visible from the guard discarded, as illustrated in Fig.~\ref{fig:visibility-models}.

\section{Proposed Solution for Combining and Refining Guard Sets}
\label{sec:proposed-solution}

\subsection{Hybrid Refinement Framework for the SPP}
\label{subsec:hr}

\begin{algorithm}[t]
    \caption{Hybrid Refinement (HR) Framework for the SPP}
    \label{alg:hr}
    \begin{algorithmic}[1]%
        \Statex\textbf{Input:} Environment $\mathcal{W} \cop{\subset} \mathbb{R}^2$; visibility model $\mathrm{Vis} \cop{:} \mathcal{W} \cop{\mapsto} 2^\mathcal{W}$; allowed uncovered area ratio $\epsilon \cop{\in} [0\cop{,}1]$.%
        \Statex\textbf{Parameter:} Set of SPP methods $M$, each taking the input $(\mathcal{W}, \mathrm{Vis}, \epsilon)$ and returning a finite guard set $G \cop{\subset} \mathcal{W}$.%
        \Statex\textbf{Output:} Refined finite guard set $G \cop{\subset} \mathcal{W}$.
        \State $G \gets \emptyset$\label{alg:hr:l1}%
        \For{$\mathrm{Method} \in M$}
            \label{alg:hr:l2}%
            \State $G \gets G \cup \mathrm{Method}(\mathcal{W}, \mathrm{Vis}, \epsilon)$\label{alg:hr:l3}%
        \EndFor%
        \State $C \gets \{(g, \mathcal{V}) \mid g \in G \land \mathcal{V} = \mathrm{Vis}(g) \}$\label{alg:hr:l4}%
        \State $G \gets{}$\Call{RefineCoverage}{$C$, $\mathcal{W}$, $\epsilon$}\label{alg:hr:l5}%
        \State \Return $G$\label{alg:hr:l6}%
        \Function{RefineCoverage}{$C$, $\mathcal{W}$, $\epsilon$}
            \label{alg:hr:l7}%
            \State $(g_1, \mathcal{V}_1) \ldots (g_n, \mathcal{V}_n) \gets C$\label{alg:hr:l8}%
            \State $G \gets \emptyset$; $L \gets \{1, \ldots, n\}$; $\mathrm{cov} \gets 0$\label{alg:hr:l9}%
            \While{$L \neq \emptyset \textbf{ and } \mathrm{cov} < (1 - \epsilon) \mathrm{Area}(\mathcal{W})$}
                \label{alg:hr:l10}%
                \State $k \gets \argmax_{\,i \in L} \mathrm{Area}(\mathcal{V}_{i})$\label{alg:hr:l11}%
                \State $\mathrm{cov} \gets \mathrm{cov} + \mathrm{Area}(\mathcal{V}_{k})$\label{alg:hr:l12}%
                \For{$i \in L$}
                    \label{alg:hr:l13}%
                    \State $\mathcal{V}_i \gets \mathrm{cl}(\mathcal{V}_i \setminus \mathcal{V}_k)$\label{alg:hr:l14}%
                    \If{$\mathcal{V}_i = \emptyset$}
                        \label{alg:hr:l15}%
                        \State $L \gets L \setminus \{i\}$\label{alg:hr:l16}%
                    \EndIf%
                \EndFor%
                \State $G \gets G \cup \{g_k\}$\label{alg:hr:l17}%
            \EndWhile
            \State \Return $G$\label{alg:hr:l18}%
        \EndFunction%
    \end{algorithmic}
\end{algorithm}

The \emph{hybrid refinement} (HR) framework takes a set of sensor-placement methods and refines their output by reducing the number of guards while maintaining sufficient coverage.
Its primary goal is to minimize redundancy while ensuring that coverage quality remains above a user-defined threshold, provided the initial guard set meets this requirement.
The HR framework applies to any SPP variant and method, though this work evaluates it specifically for the OSPP in polygonal environments.
HR processes the combined output—referred to as the \emph{initial guard set}—which may or may not meet the coverage requirement in Eq.~\eqref{eq:problem-definition}.
If it does, HR ensures that the refined guard set $G$ also meets this requirement while eliminating guards with little or no contribution, keeping coverage quality within $[1 \cop{-} \epsilon\cop{,} 1]$.
Conversely, if the initial guard set fails to meet the coverage requirement, HR preserves its coverage quality while still removing any guards that do not~contribute~at~all.

The HR framework is outlined in Alg.~\ref{alg:hr}, while its two core concepts are illustrated in Fig.~\ref{fig:main-concepts}.
The algorithm takes three main inputs: the environment $\mathcal{W}$, the visibility model $\mathrm{Vis}$, and the allowed uncovered area ratio $\epsilon$.
Additionally, HR requires a set of SPP methods, $M$, where each method takes the same inputs as HR and returns a finite guard set.
Lines~\ref{alg:hr:l1}--\ref{alg:hr:l3} generate the initial guard set $G$ using these methods.
Line~\ref{alg:hr:l4} then prepares the coverage set $C$ for refinement, performed in line~\ref{alg:hr:l5}, with the final output returned in line~\ref{alg:hr:l6}.
The \textsc{RefineCoverage} function implements the refinement step using a greedy approach that iteratively selects guards covering the largest uncovered area and eliminates redundant guards.

\textsc{RefineCoverage} takes the coverage set $C$, along with $\mathcal{W}$ and $\epsilon$, as input and outputs a refined guard set $G$.
Lines~\ref{alg:hr:l8}--\ref{alg:hr:l9} initialize $G$ as empty, set $L$ to include all initial guard IDs, and set the total covered area $\mathrm{cov}$ to zero.
Each element in $C$ is a pair $(g_i, \mathcal{V}_i)$, where $g_i$ is a guard and $\mathcal{V}_i$ is its visibility region.
During refinement, $G$ is iteratively (re)constructed, potentially excluding some initial guards.
As refinement progresses, the regions $\mathcal{V}_i$ transition into \emph{residual regions}, potentially becoming null ($\mathcal{V}_i \cop{=} \emptyset$), at which point they are \emph{always} eliminated.
The refinement loop (line~\ref{alg:hr:l10}) runs until the total covered area exceeds $(1 - \epsilon) \mathrm{Area}(\mathcal{W})$ or $L$ is empty.
In each iteration, the guard ID $k$ with the largest residual region is selected (line~\ref{alg:hr:l11}), and $\mathrm{cov}$ is updated (line~\ref{alg:hr:l12}).
Next, $k$'s residual region is subtracted from all remaining guards (line~\ref{alg:hr:l14}), eliminating those with empty residuals (line~\ref{alg:hr:l16}).
Finally, $k$ is added to $G$ (line~\ref{alg:hr:l17}), and the process continues until termination.
Note that the process can terminate before $L$ is empty, in which case the remaining guards in $L$ contribute only marginally to coverage and are thus also eliminated.

The computational efficiency of HR largely depends on the implementation of the difference operation in line~\ref{alg:hr:l14}.
Alg.~\ref{alg:hr} presents a naive implementation that lacks preprocessing, requiring the full $\mathcal{V}_k$ to be clipped from each $\mathcal{V}_i$ in every iteration.
Since this operation is repeated for all remaining residual regions, and both $\mathcal{V}_i$ and $\mathcal{V}_k$ can be highly complex, it risks becoming the bottleneck of HR\@.
To mitigate this, we introduce an acceleration technique to preprocess residual regions, significantly accelerating the difference operation and enhancing overall performance.

\subsection{Accelerated Refinement Procedure}
\label{subsec:ar}

\begin{algorithm}[t!]
    \caption{Accelerated Refinement Procedure}
    \label{alg:ar}
    \begin{algorithmic}[1]
        \Statex\textbf{Additional parameter:} Grid cell size $s \cop{\in} \mathbb{R}^{+}$ for bucketing.
        \Function{RefineCoverage}{$C$, $\mathcal{W}$, $\epsilon$}
            \label{alg:ar:l1}
            \State $(g_1, \mathcal{V}_1) \ldots (g_n, \mathcal{V}_n) \gets C$ \label{alg:ar:l2}
            \State $B_1 \ldots B_n \gets{}$\Call{Preprocess}{$\mathcal{V}_1 \ldots \mathcal{V}_n$, $\mathcal{W}$, $s$} \label{alg:ar:l3}
            \State $G \gets \emptyset$; $L \gets \{1, \ldots, n\}$; $\mathrm{cov} \gets 0$ \label{alg:ar:l4}
            \While{$L \neq \emptyset \textbf{ and } \mathrm{cov} < (1 - \epsilon) \mathrm{Area}(\mathcal{W})$} \label{alg:ar:l5}
            \State $k \gets \argmax_{\,i \in L} \sum_{\mathcal{B}\in B_i}\mathrm{Area}(\mathcal{B})$ \label{alg:ar:l6}
            \State $\mathrm{cov} \gets \mathrm{cov} + \sum_{\mathcal{B}\in B_k}\mathrm{Area}(\mathcal{B})$ \label{alg:ar:l7}
            \For{$i \in L$} \label{alg:ar:l8}
            \State $B_i \gets{}$\Call{Difference}{$B_i$, $B_k$} \label{alg:ar:l9}
            \If{$B_i = \emptyset$} \label{alg:ar:l10}
            \State $L \gets L \setminus \{i\}$ \label{alg:ar:l11}
            \EndIf
            \EndFor
            \State $G \gets G \cup \{g_k\}$ \label{alg:ar:l12}
            \EndWhile
            \State \Return $G$ \label{alg:ar:l13}
        \EndFunction
    \end{algorithmic}
\end{algorithm}

\begin{algorithm}[t!]
    \caption{Preprocessing for Accelerated Refinement}
    \label{alg:ar-preprocess}
    \begin{algorithmic}[1]
        \Function{Preprocess}{$\mathcal{V}_1 \ldots \mathcal{V}_n$, $\mathcal{W}$, $s$} \label{alg:ar:l14}
        \State $(\mathcal{C}_{11} \ldots \mathcal{C}_{1w}) \ldots (\mathcal{C}_{h1} \ldots \mathcal{C}_{hw}) \gets \mathrm{Grid}(\mathcal{W}, s)$ \label{alg:ar:l15}
        \For{$i \gets 1 \ldots  n$} \label{alg:ar:l16}
        \State $B_i \gets \emptyset$ \label{alg:ar:l17}
        \For{$x \gets 1 \ldots  w$} \label{alg:ar:l18}
        \For{$y \gets 1 \ldots  h$} \label{alg:ar:l19}
        \State $\mathcal{B} \gets \mathcal{V}_i \cap \mathcal{C}_{xy}$ \label{alg:ar:l20}
        \If{$\mathcal{B} \neq \emptyset$} \label{alg:ar:l21}
        \State $B_i \gets B_i \cup \{\mathcal{B}\}$ \label{alg:ar:l22}
        \EndIf
        \EndFor
        \EndFor
        \EndFor
        \State \Return $B_1 \ldots B_n$ \label{alg:ar:l23}
        \EndFunction
    \end{algorithmic}
\end{algorithm}

\begin{algorithm}[t!]
    \caption{Difference Operation for Residual Regions}
    \label{alg:ar-difference}
    \begin{algorithmic}[1]
        \Function{Difference}{$B$, $K$} \label{alg:ar:l24}
        \State $D \gets \emptyset$ \label{alg:ar:l25}
        \For {$\mathcal{B} \in B$} \label{alg:ar:l26}
        \For {$\mathcal{K} \in K$} \label{alg:ar:l27}
        \If{\Call{CanIntersect}{$\mathcal{B}$, $\mathcal{K}$}} \label{alg:ar:l28}
        \State $\mathcal{D} \gets \mathrm{cl}(\mathcal{B} \setminus \mathcal{K})$ \label{alg:ar:l29}
        \If{$\mathcal{D} \neq \emptyset$} \label{alg:ar:l30}
        \State $D \gets D \cup \{\mathcal{D}\}$ \label{alg:ar:l31}
        \EndIf
        \Else \label{alg:ar:l32}
        \State $D \gets D \cup \{\mathcal{B}\}$ \label{alg:ar:l33}
        \EndIf
        \EndFor
        \EndFor
        \State \Return $D$ \label{alg:ar:l34}
        \EndFunction
    \end{algorithmic}
\end{algorithm}

\begin{algorithm}[t!]
    \caption{Intersection Check for Residual Regions}
    \label{alg:ar-canintersect}
    \begin{algorithmic}[1]
        \Function{CanIntersect}{$\mathcal{B}$, $\mathcal{K}$} \label{alg:ar:l35}
        \If{$\mathrm{GridCell}(\mathcal{B}) = \mathrm{GridCell}(\mathcal{K})$} \label{alg:ar:l36}
        \If{$\mathrm{BoundBox}(\mathcal{B}) \cap \mathrm{BoundBox}(\mathcal{K}) \neq \emptyset$} \label{alg:ar:l37}
        \If{$\mathrm{BoundBox}(\mathcal{B}) \cap \mathcal{K} \neq \emptyset$} \label{alg:ar:l38}
        \State \Return \textbf{true} \label{alg:ar:l39}
        \EndIf
        \EndIf
        \EndIf
        \State \Return \textbf{false} \label{alg:ar:l40}
        \EndFunction
    \end{algorithmic}
\end{algorithm}

To overcome the inefficiency of the naive implementation, we introduce a faster approach leveraging an acceleration technique akin to the well-known \emph{bucketing technique} for the point-location problem~\citep{Edahiro1984}.
The accelerated refinement procedure is detailed in Algs.~\ref{alg:ar}--\ref{alg:ar-canintersect}.
This technique partitions the environment $\mathcal{W}$ into a grid of square cells with side length $s$, which is a new parameter.
Each visibility region $\mathcal{V}_i$ is then preprocessed into a set $B_i$ of smaller regions by intersecting $\mathcal{V}_i$ with grid cells and storing only non-empty intersections (line~\ref{alg:ar:l3} of Alg.~\ref{alg:ar}, Alg.~\ref{alg:ar-preprocess}).
The difference operation then processes these regions, which have significantly reduced size and complexity compared to the original visibility regions (line~\ref{alg:ar:l9} of Alg.~\ref{alg:ar}, Alg.~\ref{alg:ar-difference}).
Additionally, the proposed fast difference operation eliminates unnecessary computations by first checking whether $\mathcal{B} \cop{\in} B_i$ and $\mathcal{K} \cop{\in} B_k$ can intersect (line~\ref{alg:ar:l28} of Alg.~\ref{alg:ar-difference}).
Only if an intersection is possible does the algorithm perform the clipping operation (line~\ref{alg:ar:l29} of Alg.~\ref{alg:ar-difference}).
Otherwise, $\mathcal{B}$ is directly added to the difference set $D$ (line~\ref{alg:ar:l33} of Alg.~\ref{alg:ar-difference}).

The \textsc{CanIntersect} function (line~\ref{alg:ar:l28} of Alg.~\ref{alg:ar-difference}, Alg.~\ref{alg:ar-canintersect}) efficiently determines whether two regions can intersect by verifying three conditions: whether they belong to the same grid cell (which can be tabulated), whether their bounding boxes overlap (a fast check), and whether the first region’s bounding box intersects the second region (which remains significantly faster than computing the full difference).
This function provides a necessary but not sufficient condition for intersection, so the two regions may still not intersect even if this function returns true.
However, this is not an issue, as its primary role is to quickly eliminate non-intersecting regions and avoid unnecessary expensive difference operations.

\section{Related Work}
\label{sec:related-work}

The original SPP, known as the \emph{art gallery problem} (AGP)~\citep{Chvatal1975}, is a fundamental problem in \emph{computational geometry}.
Like our formulation, the AGP seeks to minimize the number of guards required to cover an environment.
However, it differs in three key aspects: it assumes a simple polygon as the environment (ours considers holes), uses an unlimited omnidirectional visibility model (ours is more general), and enforces complete coverage (ours allows a user-defined coverage ratio).
Chvátal's theorem~\citep{Chvatal1975} established that $\lfloor n/3 \rfloor$ guards are sometimes necessary and always sufficient to guard a simple polygon with $n$ vertices.
Since then, extensive research has produced numerous theoretical results, enough to fill a book~\citep{ORourke1987}.
For instance, it has been shown that the AGP in polygons with holes is both NP-hard~\citep{ORourke1983,Lee1986} and APX-hard~\citep{Eidenbenz2001}.

In practical applications such as robotic inspection and security surveillance, various visibility constraints must be considered.
These may include limited visibility range, localization uncertainty (as in our case), or additional factors like a maximum angle of incidence, further complicating the problem.
The challenge is amplified in large, complex environments where exact solutions become impractical.
Consequently, SPP applications typically rely on heuristic methods, which we discuss next.

In robotics, González-Baños and Latombe conducted influential research on a variant of the SPP~\citep{Gonzalez-Banos1998,Gonzalez-Banos2001}, proposing two randomized sampling methods for minimizing the number of guards needed to cover a polygonal environment's boundary.
Their approach incorporated constraints on minimum and maximum visibility range, as well as the maximum angle of incidence.
While originally designed for boundary coverage with additional constraints, these methods can be readily adapted to cover the entire environment and address the OSPP formulation presented in this paper.

The first sampling approach involves extensive random sampling of the environment until complete boundary coverage is achieved.
The next step is to identify the smallest subset of these samples that maintains full coverage, a process similar to our refinement procedure.
To do this, the boundary is divided into maximal connected components, each fully covered by a subset of guards.
These components are assigned unique labels, forming the \emph{universe} $U$.
A family $S$ of subsets of $U$ is then constructed, where each subset corresponds to a single guard and contains the component labels it covers.
The objective is to find the smallest subfamily $C \cop{\subset} S$ whose union equals $U$, a problem known as the \emph{minimum set cover problem} (MSCP), one of Karp's original NP-hard problems~\citep{Karp1972}.
Beyond being NP-hard, the MSCP is also APX-hard~\citep{Feige1998}.
Despite this complexity, González-Baños and Latombe proposed a greedy heuristic that, in each iteration, selects the subset covering the largest number of yet-uncovered components.

González-Baños and Latombe identified two main drawbacks of the previous approach that may limit its practical usability: a quadratic dependency on the number of samples and the redundancy of many samples due to unsystematic random sampling.
To mitigate these issues, they proposed an alternative \emph{incremental} method incorporating the \emph{dual sampling} (DS) scheme.
Instead of sampling the entire environment, this method selectively samples constraint points on the boundary.
The DS scheme iteratively selects an uncovered boundary point, determines its visible region, and places dual samples within that region.
The dual sample covering the largest portion of the uncovered boundary is added to the partial solution, and the process repeats until the desired coverage is achieved.
Notably, this approach can be extended beyond boundary coverage to cover entire environments, aligning with the focus of this work.

Beyond sampling-based heuristics, several \emph{convex-partitioning} methods have been proposed for the OSPP\@.
These approaches partition the environment into disjoint convex regions (disjoint except at boundaries), each fully coverable by a single guard.
One such method is \emph{constrained conforming Delaunay triangulation} (CCDT)~\citep{Shewchuk2002}, which generates a triangular mesh of the polygonal environment using a refinement procedure based on the \emph{Delaunay triangulation}.
In CCDT, the mesh is constrained by the environment’s boundaries and conforms to user-defined constraints---specifically, maximum circumcircle radii for the limited-range visibility model.
For the unlimited visibility model, no additional constraints are imposed, yielding the \emph{constrained Delaunay triangulation} (CDT).
Guards are then placed at the circumcenters of acute triangles and at the midpoints of the bases of obtuse triangles.

Kazazakis and Argyros~\citep{Kazazakis2002} proposed a heuristic specifically for the limited-range visibility model.
This iterative approach starts with a convex polygonal mesh of the environment, which can be obtained by merging triangles from the CDT\@.
The method then progressively subdivides convex polygons, refining the mesh to satisfy the limited-range visibility constraint.
Guards are placed within each resulting polygon at a weighted average of the edge midpoints, where the weights correspond to edge lengths.

The last most relevant method falls outside the previously discussed categories of sampling and convex-partitioning techniques.
It involves placing guards at all reflex vertices of the environment (or a single guard at any location if the environment is convex).
This approach, proposed in~\citep{Sarmiento2004} for robotic search tasks, is supposed to ensure complete coverage under the unlimited visibility model.
However, we have found no formal proof that this guarantee indeed holds for polygons with holes.
There is a result for simple polygons stating that placing guards at all reflex vertices guarantees complete coverage~\citep{Urrutia2000}.
Additionally,~\citep{ORourke1987} implies that placing guards at all convex vertices guarantees coverage of a simple polygon's exterior.
A proof of the method's sufficiency for polygons with holes could build on these results, showing that adding a hole with guards at its convex vertices (which are reflex in the polygon with the hole) does not reduce coverage.
While we conjecture this to be true, formalizing the proof is beyond the scope of this paper.
For now, we confirm that this method works for all evaluated instances.

We now position our proposed HR framework within the context of the previously discussed approaches.
The framework revisits the idea of transforming the SPP into the MSCP and solving it via a greedy algorithm but differs significantly from the first method of González-Baños and Latombe.
First, instead of random sampling, it systematically generates an initial guard set using fast SPP methods that collectively ensure coverage of the desired quality.
As demonstrated in our computational study, incorporating diverse initial methods leads to a smaller guard set after refinement, a key insight of the HR framework.
Second, unlike traditional MSCP approaches, the HR framework does not explicitly construct the universe $U$ and the family of subsets $S$.
Instead, the greedy refinement algorithm selects guards based on the maximum uncovered area they cover, aligning more closely with the incremental approach of the DS scheme than with the maximum-cardinality strategy of MSCP\@.
However, the HR framework further diverges from the DS scheme by forgoing dual sampling; instead, the set of candidate samples is predetermined by the initial SPP methods, enabling preprocessing and acceleration techniques such as bucketing, yielding the final \emph{hybrid accelerated-refinement} (HAR) framework.
Moreover, convex-partitioning and reflex-vertex methods are well-suited as initial SPP methods in HR, given their speed and guaranteed coverage for certain visibility models.
In conclusion, the HAR framework integrates key ideas from all discussed approaches into a unique solution strategy, enhanced by an acceleration technique.

Finally, we discuss the possibility of extending the HR framework with a more sophisticated refinement step.
While the MSCP solution strategy is not limited to greedy algorithms, their simplicity and efficiency make them highly practical.
As shown in our computational study, the greedy approach within the HR framework achieves the best trade-off between computational time and solution quality compared to the discussed baselines.
However, alternative refinement strategies remain worth exploring.
Although beyond the scope of this paper, recent metaheuristic approaches for the MSCP, surveyed in~\citep{Rosenbauer2020}, could potentially be adapted to the SPP and integrated into the HAR framework.
We leave this as a promising direction for future research.

\section{Large-Scale Computational Study}
\label{sec:computational-evaluation}

\subsection{Evaluation Metrics}
\label{subsec:evaluation-metrics}

The performance of any SPP method can be assessed using three key metrics: the number of returned guards, $n \cop{\coloneqq} |G|$, the measured \emph{runtime} $t$, and the \emph{coverage ratio}:
\begin{equation}
    \%\mathrm{CR} \coloneqq \frac{\mathrm{Area}\left(\bigcup_{g \in G} \mathrm{Vis}(g)\right)}{\mathrm{Area}(\mathcal{W})} \cdot 100\%.
    \label{eq:coverage-ratio}
\end{equation}
However, if a method \emph{guarantees} complete coverage or at least $(1 - \epsilon)$ coverage, $\%\mathrm{CR}$ becomes redundant and is omitted unless explicitly stated.

Comparing methods solely by the number of guards $n$ is challenging due to high variance across problem instances.
Thus, we introduce the \emph{percentage best-known solution gap}, $\%\mathrm{BG}(n) \cop{\in} [0, \infty)$, which quantifies the relative difference between the number of guards returned by a method and the best-known solution $n_{\mathit{best}}$ for the given instance:
\begin{equation}
    \%\mathrm{BG}(n)  \coloneqq  \frac{n - n_{\mathit{best}}}{n_{\mathit{best}}} \cdot 100\%.
    \label{eq:best-known-gap}
\end{equation}
Here, $n_{\mathit{best}}$ is the minimum number of guards recorded across all evaluated methods and runs for a given instance, with non-deterministic methods run multiple times to account for randomness.
Crucially, this gap is computed relative to the \emph{best-known solution} obtained from the evaluated heuristic methods rather than the \emph{optimal solution}, which remains unknown for large-scale instances due to their intractability.

\subsection{Benchmark Instances}
\label{subsec:benchmark-instances}

Our problem setup consists of the polygonal environment $\mathcal{W}$, the visibility model $\mathrm{Vis}$, and the coverage parameter $\epsilon$, fixed at $0.001$ for all experiments.
The environments are derived from a dataset of 35 polygonal maps from the videogame Iron Harvest, developed by KING Art Games, as introduced in~\citep{Harabor2022}.
These maps, typically spanning $400 \cop{\times} 400$ meters, feature thousands of vertices and dozens to hundreds of holes, making them ideal for evaluating large-scale, complex scenarios.
To ensure well-formed environments, we preprocess the maps to create connected polygons with holes, eliminating self-intersections, overlapping holes, and redundant vertices.
This involves a sequence of smoothing operations: applying the \emph{Ramer-Douglas-Peucker} (RDP) algorithm~\citep{Douglas1973} with a $0.1\,\mathrm{m}$ tolerance, performing inflation–deflation with a $0.2\,\mathrm{m}$ radius, a final inflation of $0.01\,\mathrm{m}$, and reapplying RDP with a $0.1\,\mathrm{m}$ tolerance.
The preprocessing concludes by selecting the largest polygon from the refined representation, preserving all enclosed holes, and discarding any disconnected artifacts such as isolated or polygons fully enclosed in holes.
The properties of the resulting maps are summarized in Tab.~\ref{tab:map-properties}.

\begin{table}[t!]
    \centering
    \begin{threeparttable}
        \caption{Map Properties.}
        \label{tab:map-properties}
        \scriptsize
        \begin{tabular}{*{7}{r}c}
            \toprule
            Map     & $n$  & $h$ & $x$ & $y$ & $xy$    & $a$     & Prelim. Eval. \\
            \midrule
            2p01    & 1909 & 140 & 189 & 210 & 39,606  & 31,484  &            \\
            2p02    & 1428 & 137 & 270 & 270 & 72,889  & 53,909  &            \\
            2p03    & 2347 & 153 & 330 & 310 & 102,300 & 58,924  & \checkmark \\
            2p04    & 998  & 52  & 240 & 310 & 74,389  & 52,758  &            \\
            4p01    & 2919 & 274 & 320 & 320 & 102,387 & 75,312  &            \\
            4p02    & 3799 & 315 & 380 & 502 & 190,605 & 109,607 & \checkmark \\
            4p03    & 4838 & 300 & 400 & 410 & 163,984 & 97,727  &            \\
            6p01    & 3558 & 234 & 368 & 498 & 183,584 & 121,713 &            \\
            6p02    & 3419 & 214 & 400 & 440 & 175,792 & 130,409 & \checkmark \\
            6p03    & 2464 & 229 & 500 & 500 & 249,980 & 151,745 &            \\
            cha01   & 1357 & 112 & 230 & 280 & 64,390  & 41,997  & \checkmark \\
            cha02   & 2108 & 101 & 335 & 570 & 190,932 & 176,227 &            \\
            cha03   & 3462 & 320 & 400 & 430 & 171,983 & 99,653  &            \\
            cha04   & 4688 & 407 & 440 & 440 & 193,582 & 120,911 &            \\
            endmaps & 4923 & 340 & 565 & 770 & 435,023 & 360,408 &            \\
            pol01   & 959  & 51  & 323 & 133 & 42,798  & 12,839  &            \\
            pol02   & 3296 & 239 & 470 & 515 & 242,030 & 96,545  &            \\
            pol03   & 4118 & 394 & 420 & 510 & 214,181 & 127,080 &            \\
            pol04   & 3978 & 268 & 350 & 340 & 118,986 & 72,535  & \checkmark \\
            pol05   & 2860 & 239 & 515 & 395 & 203,198 & 85,389  &            \\
            pol06   & 5315 & 465 & 470 & 480 & 225,581 & 156,673 & \checkmark \\
            rus01   & 2331 & 134 & 331 & 224 & 73,976  & 33,160  &            \\
            rus02   & 1337 & 72  & 242 & 307 & 74,455  & 31,114  &            \\
            rus03   & 3463 & 295 & 450 & 430 & 193,482 & 69,337  & \checkmark \\
            rus04   & 3198 & 265 & 338 & 500 & 169,100 & 104,296 &            \\
            rus05   & 3459 & 220 & 404 & 419 & 169,285 & 84,142  & \checkmark \\
            rus06   & 5145 & 383 & 545 & 455 & 247,955 & 111,825 &            \\
            rus07   & 2147 & 137 & 460 & 380 & 174,783 & 85,189  &            \\
            sax01   & 1583 & 127 & 380 & 485 & 184,283 & 78,603  &            \\
            sax02   & 4448 & 255 & 403 & 634 & 255,845 & 117,736 & \checkmark \\
            sax03   & 2827 & 143 & 416 & 462 & 192,381 & 86,305  &            \\
            sax04   & 4639 & 286 & 585 & 675 & 394,850 & 139,834 &            \\
            sax05   & 1623 & 54  & 445 & 420 & 186,883 & 86,215  &            \\
            sax06   & 2524 & 163 & 405 & 465 & 188,308 & 96,977  &            \\
            sax07   & 2758 & 165 & 310 & 340 & 105,387 & 69,180  & \checkmark \\
            \bottomrule
        \end{tabular}
        \begin{tablenotes}
            Legend: $n$: number of vertices; $h$: number of holes; $x$: map width; $y$: map height; $xy$: bounding box area; $a$: map area, Prelim. Eval.: \checkmark{} if used only in the preliminary evaluation.
        \end{tablenotes}
    \end{threeparttable}
\end{table}

The visibility models evaluated in our study include the unlimited visibility model $\mathrm{Vis}_\infty$, the range-limited visibility model $\mathrm{Vis}_d$ with $d \cop{\in} \{4\cop{,} 6\cop{,} 8\cop{,} 12\cop{,} 16\cop{,} 24\cop{,} 32\cop{,} 48\cop{,} 64\cop{,} 96\cop{,} 128\}\,\mathrm{m}$, and the localization-uncertainty visibility model $\mathrm{Vis}_{d, \mathrm{Unc}_r}$ with $r \cop{\in} \{0.1\cop{,} 0.2\cop{,} 0.4\cop{,} 0.8\cop{,} 1.6\}\,\mathrm{m}$, using the same $d$ values as $\mathrm{Vis}_d$.
To approximate visibility regions, the sampling parameter $d_{\text{samp}}$ is set such that for the smallest range $d \cop{=} 4$, the open-space visibility region is represented by a 12-sided regular polygon: $d_{\text{samp}} \cop{=} \frac{4 \cdot 2\pi}{12} \cop{\approx} 2.094\,\mathrm{m}$.
Similarly, for localization uncertainty, $r_{\text{samp}}$ is chosen so that the smallest nonzero uncertainty radius, $r \cop{=} 0.1$, is approximated with at least four samples plus the guard itself: $r_{\text{samp}} \cop{=} \frac{0.1 \cdot 2\pi}{4} \cop{\approx} 0.157\,\mathrm{m}$.
For details on these models and parameters, refer to Sec.~\ref{subsec:visibility-models}.

\begin{figure}[t!]
    \centering
    \begin{subfigure}[t]{0.32\columnwidth}
        \centering
        \includegraphics[width=\linewidth]{g/vis_inf-64-32-16-8-4_rob_0}
        \caption*{$\mathrm{Vis}_\infty$, $\mathrm{Vis}_d$}
        \caption*{($d\cop{=}64, 32, 16, 8, 4\,\mathrm{m}$)}
    \end{subfigure}
    \begin{subfigure}[t]{0.32\columnwidth}
        \centering
        \includegraphics[width=\linewidth]{g/vis_inf_rob_0-10-20-40-80-160}
        \caption*{$\mathrm{Vis}_\infty$, $\mathrm{Vis}_{\infty, \mathrm{Unc}_r}$}
        \caption*{($r\cop{=}0.1, 0.2, 0.4, 0.8, 1.6\,\mathrm{m}$)}
    \end{subfigure}
    \begin{subfigure}[t]{0.32\columnwidth}
        \centering
        \includegraphics[width=\linewidth]{g/vis_16_rob_0-10-20-40-80-160}
        \caption*{$\mathrm{Vis}_{d=16}$, $\mathrm{Vis}_{d=16, \mathrm{Unc}_r}$}
        \caption*{($r\cop{=}0.1, 0.2, 0.4, 0.8, 1.6\,\mathrm{m}$)}
    \end{subfigure}
    \caption{
        Showcase of the visibility models used in the study on the 2p01 map.
        The guard location is marked by a yellow central dot in all images.
        The left image displays the unlimited visibility model $\mathrm{Vis}_\infty$ and the range-limited model $\mathrm{Vis}_{d}$, stacked in sequence from blue ($\mathrm{Vis}_\infty$) to red ($\mathrm{Vis}_{d=4}$).
        The middle and right images depict the localization-uncertainty models $\mathrm{Vis}_{\infty, \mathrm{Unc}_r}$ and $\mathrm{Vis}_{d=16, \mathrm{Unc}_r}$, with colors transitioning from blue ($r \cop{=} 0$) to cyan ($r \cop{=} 1.6\,\mathrm{m}$) and from purple ($r \cop{=} 0$) to green ($r \cop{=} 1.6\,\mathrm{m}$), respectively.
    }
    \label{fig:2p01-models}
\end{figure}

\begin{figure}[t!]
    \centering
    \begin{subfigure}[t]{0.32\columnwidth}
        \centering
        \includegraphics[trim={40cm, 55cm, 40cm, 55cm}, clip, width=\linewidth]{g/vis_16_rob_0-10-20-40-80-160}
    \end{subfigure}
    \hfill
    \begin{subfigure}[t]{0.32\columnwidth}
        \centering
        \includegraphics[trim={4mm, 4mm, 4mm, 4mm}, clip, width=\linewidth]{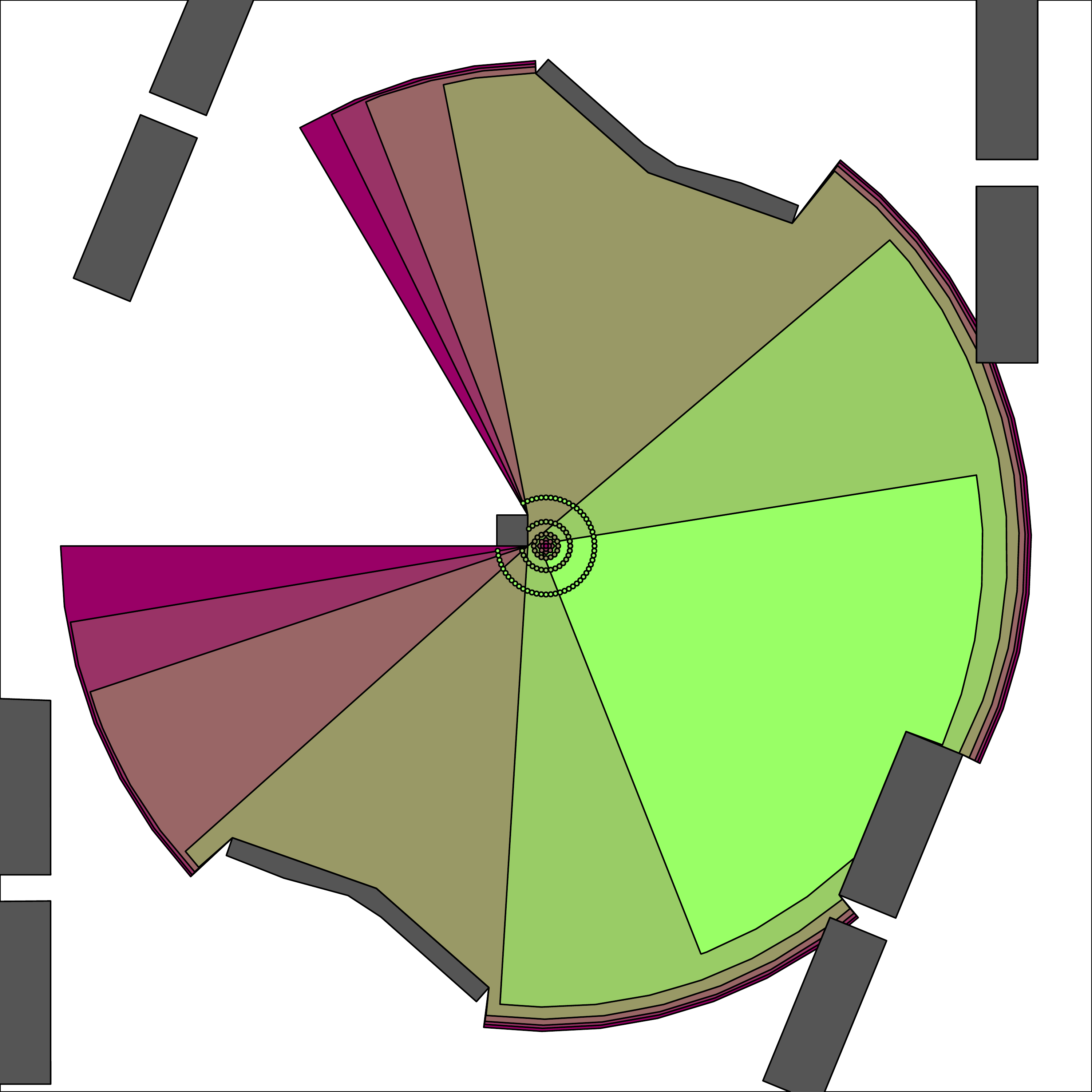}
    \end{subfigure}
    \hfill
    \begin{subfigure}[t]{0.32\columnwidth}
        \centering
        \includegraphics[trim={4mm, 4mm, 4mm, 4mm}, clip, width=\linewidth]{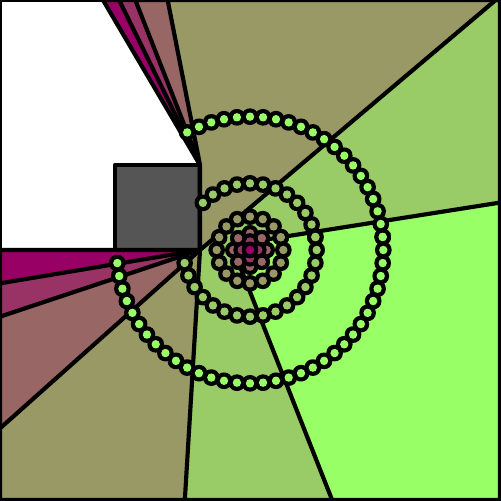}
    \end{subfigure}
    \caption{
        Close-up views of $\mathrm{Vis}_{d=16}$ and $\mathrm{Vis}_{d=16, \mathrm{Unc}_r}$.
        The middle and right images show the samples used to approximate the visibility regions.
        In $\mathrm{Vis}_{d=16}$ (purple), a single sample represents the guard.
        For uncertainty models, additional samples form concentric rings at distance $r$, transitioning from purple to green as $r$ increases.
    }
    \label{fig:2p01-models-detail}
\end{figure}

To illustrate the scale and complexity of the benchmark instances and to validate the chosen visibility model parameters, we present visualizations of the 2p01 map and its visibility models in Fig.~\ref{fig:2p01-models}, with a detailed view in Fig.~\ref{fig:2p01-models-detail}.
This selected map is relatively small and moderately complex, featuring 1,909 vertices, 140 holes, and a total area of $31{,}484\,\mathrm{m}^2$.
For comparison, the largest map by area, endmaps, contains 4,923 vertices, 340 holes, and spans $360{,}408\,\mathrm{m}^2$, while the most complex map by vertex count, pol06, has 5,315 vertices, 465 holes, and covers $156{,}673\,\mathrm{m}^2$.

\subsection{Evaluated Methods}
\label{subsec:evaluated-methods}

We evaluate a range of heuristic methods for the OSPP, including the following baselines:
\begin{enumerate*}[label=(\arabic*)]
    \item RV~\citep{Sarmiento2004}: Places guards at \emph{reflex vertices} of $\mathcal{W}$, ensuring complete coverage under $\mathrm{Vis}_\infty$.
    \item CCDT~\citep{Shewchuk2002}: A convex-partitioning method for $\mathrm{Vis}_d$ that constructs a CCDT mesh, placing guards at circumcenters of acute triangles and base midpoints of obtuse ones.
    \item KA~\citep{Kazazakis2002}: Another convex-partitioning approach tailored for $\mathrm{Vis}_d$.
    \item RS: A naive \emph{random sampling} method that places guards at uniformly sampled points in $\mathcal{W}$ until the coverage criterion is met.
    \item IRS: An \emph{informed} sampling method that dynamically updates the uncovered region, sampling exclusively from this region to improve efficiency.
    \item DSk~\citep{Gonzalez-Banos2001}: A dual sampling method where a guard is first placed at a uniform sample from the uncovered region.
    Then, $k$ additional samples are drawn from its visibility region, selecting the one maximizing coverage.
    \item DS$\uprho$: A density-based variant of DS, where the number of dual samples $k$ is set proportionally to the visibility region area, $k \cop{=} \lceil \rho \mathrm{Area}(\mathcal{V}) \rceil$.
\end{enumerate*}
For further background, see Sec.~\ref{sec:related-work}.
DSk is evaluated with $k \cop{\in} \{2\cop{,}4\cop{,}8\cop{,}16\cop{,}32\cop{,}64\cop{,}128\cop{,}256\}$, while DS$\uprho$ uses $\rho \cop{\in} \{0.02\cop{,}0.04\cop{,}0.08\cop{,}0.16\cop{,}0.32\cop{,}0.64\cop{,}1.28\cop{,}2.56\}$.
The remaining baselines are parameter-free.

The proposed framework is evaluated in two variants: the naive (HR) and the accelerated (HAR).
It forms actual solution heuristics by selecting a set of guard methods, $M$.
For the unlimited visibility model $\mathrm{Vis}_\infty$, we consider 7 variants of $M$ defined by $2^A \cop{\setminus} \{\emptyset\}$, where $A \cop{=} \{\text{KA}\cop{,} \text{CCDT}\cop{,} \text{RV}\}$—three fast baseline methods that guarantee complete coverage without relying on $\epsilon$, enabling the refinement step to remove guards with minimal coverage contribution.
For the range-limited model $\mathrm{Vis}_d$, we use the same options except for $M \cop{=} \{\text{RV}\}$, as RV alone does not ensure full coverage.
Additionally, HAR is parameterized by the bucketing cell size $s$, set as $s \cop{=} \frac{1}{10} \max(x, y)$, where $x$ and $y$ are the width and height of $\mathcal{W}$, respectively.

\subsection{Implementation Details}
\label{subsec:implementation-details}

The proposed and baseline methods were implemented in C\texttt{++}, utilizing a shared codebase that includes essential components for computing triangular meshes, visibility regions, polygon clipping, and map preprocessing.

Triangular meshes are computed using Triangle\footnote{Available at \url{https://www.cs.cmu.edu/~quake/triangle.html}.}~\citep{Shewchuk1996, Shewchuk2002}, which employs \emph{Delaunay refinement algorithms}~\citep{Chew1993, Ruppert1995} and supports user-defined constraints, such as the maximum circumcircle radius.
Visibility regions are computed using the T\v{r}iVis library,\footnote{Available at \url{https://github.com/janmikulacz/trivis}.} developed by the authors~\citep{Mikula2024b, Mikula2024}, which is based on the \emph{triangular expansion algorithm}~\citep{Bungiu2014} and supports both unlimited and range-limited visibility models.
As noted in Sec.~\ref{subsec:visibility-models}, our implementation currently lacks arc-clipping support, so circular arcs are approximated with line segments.
Polygon clipping and map preprocessing operations are handled by Clipper2\footnote{Available at \url{https://github.com/AngusJohnson/Clipper2}.}~\citep{Clipper2}, which extends the \emph{Vatti clipping algorithm}~\citep{Vatti1992}.

The implementation is single-threaded, compiled in Release mode with GCC 12.3.0, and executed on a Lenovo Legion 5 Pro 16\allowbreak{}IT\allowbreak{}H6H laptop (Intel Core i7-11800H, 4.60GHz, 16GB RAM) running Ubuntu 20.04.6 LTS\@.
The full source code and reproduction scripts are available at \url{https://github.com/janmikulacz/spp}.

\subsection{Other Methodological Details}
\label{subsec:evaluation-stages}

The computational study consists of three stages, each differing in purpose and results presentation: \emph{preliminary}, \emph{main}, and \emph{additional}.

The \emph{preliminary stage} provides initial insights into the performance of all proposed and baseline methods, identifying representative methods for the main evaluation.
It focuses on the unlimited and limited-range visibility models and employs two aggregated metrics: the mean number of guards $\overline{n}$ and the mean runtime $\overline{t}$.

The \emph{main stage} delivers a more detailed evaluation of the representative methods, conducting a comprehensive comparison between the proposed and baseline heuristics.
It again focuses on the unlimited and limited-range visibility models but includes more detailed statistics, using boxplots for the number of guards $n$.
To reduce variance across different maps and visibility models, results are presented using the relative metric $\%\mathrm{BG}(n)$ (Eq.~\eqref{eq:best-known-gap}).

The \emph{additional stage} extends the analysis to the localization-uncertainty visibility model, evaluating a single version of the proposed framework.
Since the method does not guarantee coverage quality under this model, an additional metric, the percentage covered ratio $\%\mathrm{CR}$ (Eq.~\eqref{eq:coverage-ratio}), is included.

The preliminary stage evaluates all 39 proposed and baseline method variants (Sec.~\ref{subsec:evaluated-methods}), selecting 9 for the main stage.
Following best practices in computational evaluation, the benchmark instances are split into two disjoint subsets: 10 maps for the preliminary stage and 25 maps for the main and additional stages (Tab.~\ref{tab:map-properties}).
To manage experimental complexity, the preliminary stage uses a reduced set of $d$ values: $d \cop{\in} \{\infty, 64, 32, 16, 8\}\,\mathrm{m}$.

All non-deterministic methods (RS, IRS, DSk, and DS$\uprho$) are executed 10 times per instance with different random seeds to account for stochasticity.
Deterministic methods are evaluated once per instance.

\subsection{Preliminary Results for All Methods}
\label{subsec:preliminary-results}

\begin{figure}
    \centering
    \includegraphics[width=0.9\textwidth]{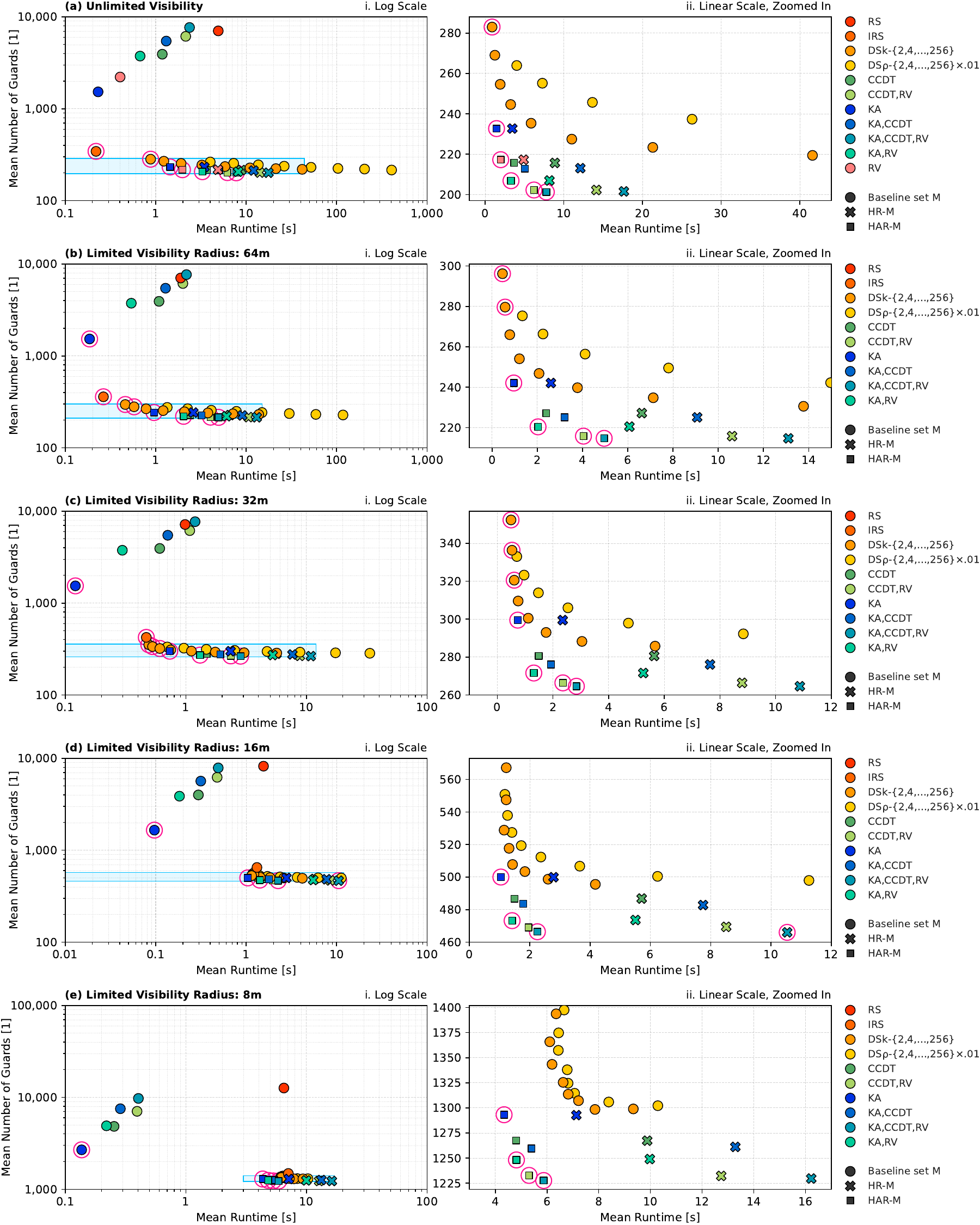}
    \caption{
        The preliminary results for five visibility models and 39 evaluated methods.
        The top row corresponds to $\mathrm{Vis}_\infty$, while the remaining rows represent range-limited models $\mathrm{Vis}_d$ with $d \cop{=} 64\cop{,} 32\cop{,} 16\cop{,} 8$.
        The left column uses a logarithmic scale, while the right column provides a zoomed-in linear-scale view of the blue-shaded region from the left column.
        Each scatterplot point represents a method as $(\overline{t}, \overline{n})$, where $\overline{t}$ is the mean runtime and $\overline{n}$ is the mean guard count, averaged over 10 maps and 10 runs for non-deterministic methods.
        Marker shapes indicate method types: circles for baselines, crosses for HR, and squares for HAR.
        Colors represent different method sets (see the legend), with non-dominated methods marked with pink circles.
    }
    \label{fig:preliminary-results}
\end{figure}

The preliminary results, shown in Fig.~\ref{fig:preliminary-results}, compare all proposed and baseline methods.
A method $a_1$ is said to dominate method $a_2$ if it achieves lower values for both aggregated metrics, $\overline{n}$ and $\overline{t}$.
Methods that are \emph{not} dominated by any other method are highlighted with pink circles.

Among baseline methods, KA, IRS, and DSk-2 are most frequently non-dominated (KA appears four times, while IRS and DSk-2 appear three times each).
KA and IRS are selected for the main evaluation---KA for its speed despite quality gaps and IRS for its more balanced performance.
DS methods are occasionally non-dominated, particularly with small dual sample counts and large visibility radii, yet they remain important baselines.
As the number of dual samples increases, solution quality improves but eventually plateaus, while computational cost rises significantly.
DSk is generally more effective than DS$\uprho$, likely because smaller visibility regions require denser sampling to minimize the number of guards.
These regions often indicate denser obstacles, leading to more varied visibility regions and necessitating finer sampling for effective coverage.
Consequently, DS$\uprho$ methods are excluded from the main evaluation.
DSk-16, DSk-64, and DSk-256 are chosen to represent moderate, high, and very high sample counts, respectively, while DS methods with fewer samples are omitted, as IRS already represents a similar approach using a single sample.

Among the proposed methods, HAR-KA, HAR-KA\cop{,}RV, and HAR-KA\cop{,}CCDT\cop{,}RV achieve non-dominated status in all five cases.
For the main evaluation, we select HAR-KA\cop{,}CCDT\cop{,}RV for its minimal guard count and HAR-KA\cop{,}RV for its strong trade-off between runtime and solution quality.
Their respective HR versions are also included to assess the impact of acceleration techniques on runtime.

\subsection{Main Results for Representative Methods}

\begin{figure}[t!]
    \centering
    \includegraphics[width=\textwidth]{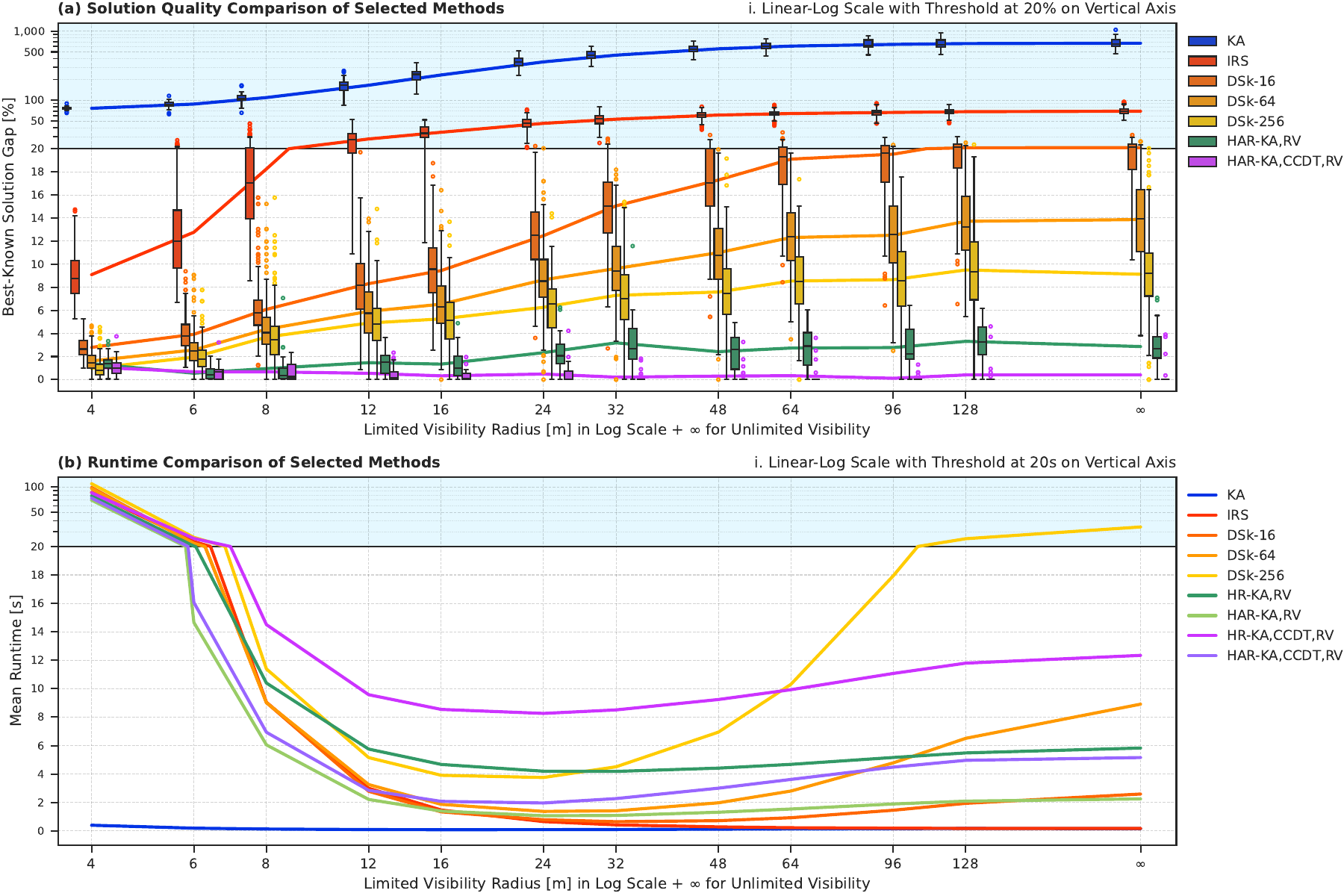}
    \caption{
        The main results for all 11 limited-range visibility models and one for $\mathrm{Vis}_\infty$, comparing the 9 representative methods.
        The top plot evaluates solution quality using the percentage best-known solution gap $\%\mathrm{BG}(n)$, while the bottom plot compares runtime $t$.
        Line plots indicate mean values for each method, while boxplots illustrate the metric distributions across 25 maps and 10 runs for non-deterministic methods.
        For clarity, runtime boxplots are omitted due to high variance caused by dataset variability.
        In the quality comparison, HAR methods are merged with their respective HR variants, as their results are statistically equivalent, aside from minor differences due to numerical inaccuracies.
        Both plots use a linear scale in the lower portion and a logarithmic scale in the upper portion, as indicated by the shaded region.
    }
    \label{fig:final-results}
\end{figure}

The main results in Fig.~\ref{fig:final-results} compare the performance of the 9 representative methods.
The KA baseline is the fastest overall, with a mean runtime $\overline{t}$ near zero across all visibility models.
However, it also yields the poorest solution quality, with $\%\mathrm{BG}(n)$ averaging 667\% for the unlimited visibility model and decreasing to 77\% at a 4\,m visibility radius.

The sampling methods aim to balance runtime and solution quality based on the number of dual samples used.
IRS, with a single sample and no dual samples, is the fastest but also has the poorest solution quality, with mean $\%\mathrm{BG}(n)$ values starting at 70\% for $\mathrm{Vis}_\infty$ and decreasing to 9\% for $d \cop{=} 4$.
DSk-16, DSk-64, and DSk-256 show mean $\%\mathrm{BG}(n)$ values of 20\%, 14\%, and 9\% for $\mathrm{Vis}_\infty$, decreasing to 3\%, 2\%, and 1\% for $d \cop{=} 4$.
As expected, increasing dual samples improves solution quality but also raises runtime.
However, the improvement plateaus, as evident from the plots, indicating diminishing returns.
Additionally, runtime reaches its minimum around $d \cop{=} 24$ and increases for both larger and smaller visibility radii.
For larger radii, the increase is due to the higher cost of computing generally larger and more complex visibility regions.
For smaller radii, it results from the need for more guards, coupled with the growing complexity of maintaining the uncovered region.
Although the uncovered region shrinks with each added guard, its complexity in terms of boundary vertex count can temporarily rise as many small regions are clipped away before ultimately reducing to near zero.

\begin{figure}[t!]
    \centering
    \includegraphics[width=0.5\textwidth]{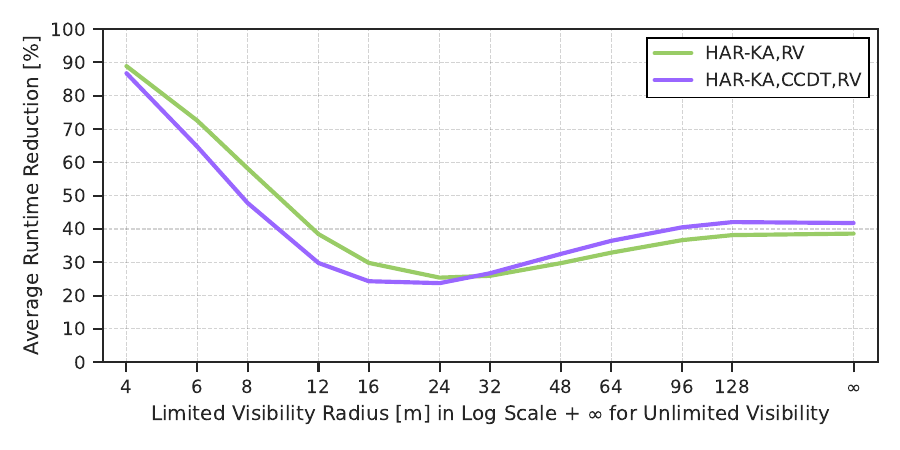}
    \caption{
        Average percentage reduction in runtime for HAR methods compared to their respective HR methods.
        In the plot, the runtime of the HR method corresponds to 100\%.
    }
    \label{fig:speedup}
\end{figure}

The proposed methods, HR and HAR, achieve the highest solution quality among all tested methods.
For the KA,CCDT,RV variant, the mean $\%\mathrm{BG}(n)$ remains near 0\% across all visibility models, while the KA,RV variant stays below 4\%.
No other methods consistently reach such low values, except for DS methods at the smallest visibility radius.
Additionally, HAR variants significantly outperform their HR counterparts in runtime.
Fig.~\ref{fig:speedup} presents the average percentage reduction in runtime for HAR compared to HR\@.
The most substantial reductions occur at $d \cop{=} 24$, with 25.4\% for KA,CCDT,RV and 23.8\% for KA,RV, corresponding to an approximate 4-fold speedup.

Overall, HAR-KA,RV offers the best balance between runtime and solution quality, consistently placing fewer guards than all baseline methods while being faster than the most competitive sampling methods, DSk-64 and DSk-256.
For $d \cop{<} 16$, it also outperforms IRS and DSk-16 in runtime, and for $d \cop{>} 16$, its mean runtime remains below 2.25 seconds.
The KA,CCDT,RV variant further reduces the guard count (by over 2\% for $d \cop{>} 24$) but incurs a slightly higher runtime, exceeding twice that of HAR-KA,RV for $d \cop{>} 24$.
For $d \cop{<} 24$, the performance gap between these variants diminishes progressively.

\subsection{Additional Results under Localization Uncertainty}

The additional results evaluate HAR-KA,RV, the best-performing method according to the main evaluation, under the localization-uncertainty visibility model $\mathrm{Vis}_{d, \mathrm{Unc}_r}$.
HR-$M$ guarantees the required coverage ratio only if the input method set $M$ ensures it.
For $\mathrm{Vis}_{d, \mathrm{Unc}_r}$, this guarantee is absent, as neither KA, RV, nor their combination ensures full coverage.
Nonetheless, we assess HAR-KA,RV's practical performance by incorporating the percentage covered ratio ($\%\mathrm{CR}$) as an additional metric.

To enhance coverage quality, we introduce minor adaptations to KA and RV for the localization-uncertainty model.
KA is adjusted to use $d \cop{-} r$ instead of $d$, aligning with the fact that, in the absence of nearby obstacles, $\mathrm{Vis}_{d, \mathrm{Unc}_r}$ forms a circle of radius $d \cop{-} r$.
RV is modified by shifting guards along the reflex vertex axis by $r \cop{+} 10^{-6}$ away from the vertex, unless this would place them outside $\mathcal{W}$.
This adjustment mitigates the visibility loss near obstacles under $\mathrm{Vis}_{d, \mathrm{Unc}_r}$, as illustrated in Fig.~\ref{fig:2p01-models-detail}, improving the likelihood that the guard remains useful in coverage and is retained during refinement.

\begin{figure}[t!]
    \centering
    \includegraphics[width=\textwidth]{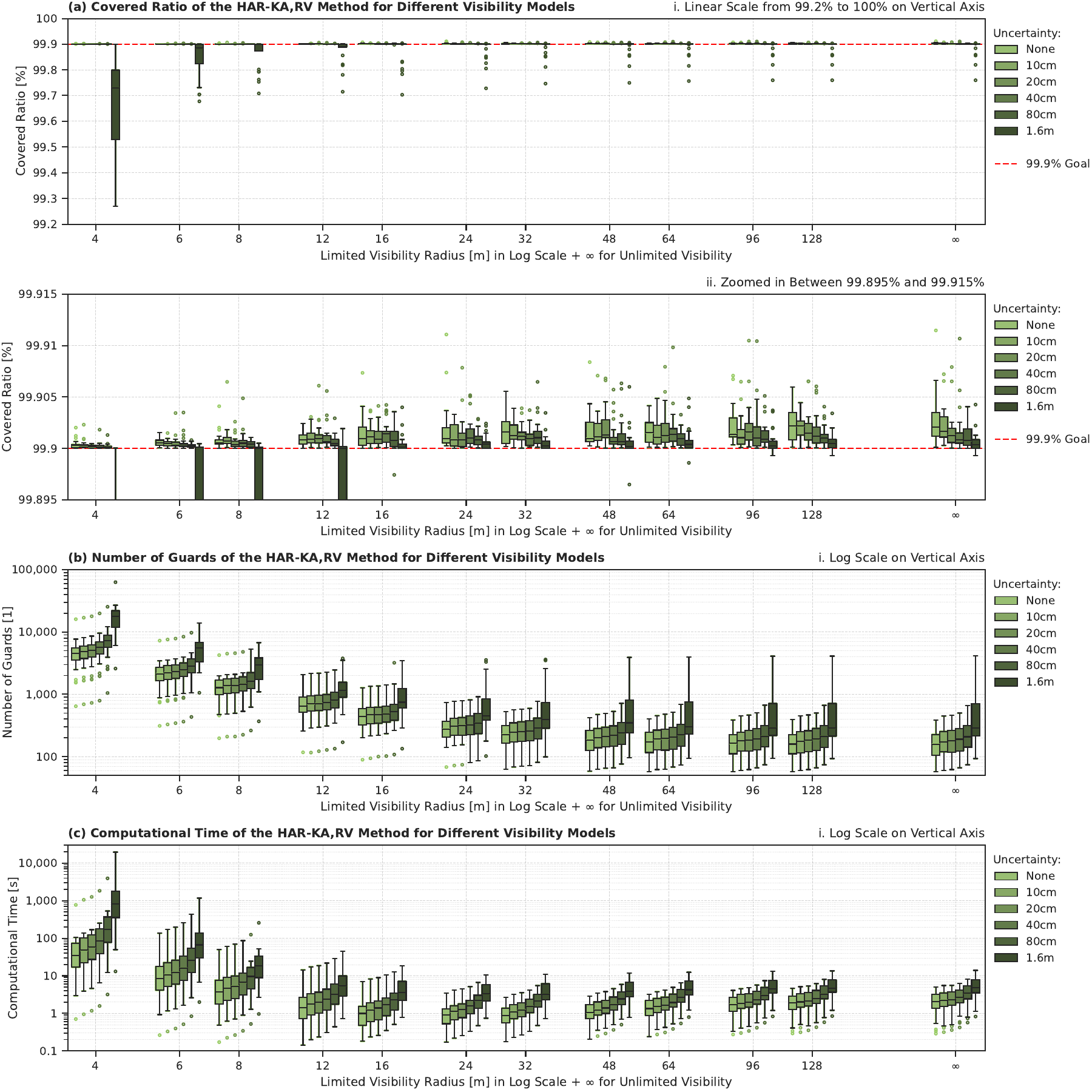}
    \caption{
        The additional results for the adapted HAR-KA,RV method under the localization-uncertainty visibility model $\mathrm{Vis}_{d, \mathrm{Unc}_r}$.
        The boxplots represent the metric distributions across all 25 maps in the main subset.
        Each boxplot corresponds to a specific visibility model, with $d$ values indicated on the horizontal axis and $r$ values distinguished by colors, as explained in the legend.
        The first plot shows the covered ratio $\%\mathrm{CR}$ within $[99.2\cop{,} 100]\%$, while the second plot zooms into $[99.895\cop{,} 99.915]\%$ for finer detail.
        A dashed red line marks the coverage ratio goal of $(1 \cop{-} \epsilon) 100\% \cop{=} 99.9\%$.
        The third and fourth plots display the number of guards $n$ and runtime $t$, respectively, both on a logarithmic scale.
    }
    \label{fig:additional-results}
\end{figure}

The results are shown in Fig.~\ref{fig:additional-results}.
Notably, the adapted HAR-KA,RV method consistently achieves the target coverage ratio of 99.9\% for $r \cop{\leq} 0.4$ across all $d$ values.
For $r \cop{=} 0.8$, coverage falls below 99.9\% in a single instance (the outlier for $d \cop{=} 16$).
For $r \cop{=} 1.6$, coverage frequently drops below 99.9\%, with the lowest recorded value at $99.2\%$ for $d \cop{=} 4$.
Thus, the adapted HAR-KA,RV method empirically meets the coverage requirement for small to moderate localization uncertainty but struggles with larger $r$ values.
The number of guards and runtime, primarily informational in this context, generally increase with larger $r$ and smaller $d$, as expected.

\section{Conclusion and Future Work}
\label{sec:conclusions}

In conclusion, this paper presents a comprehensive evaluation of heuristic methods for the OSPP in complex 2D continuous environments and introduces a novel class of HAR heuristics.
Our results demonstrate that traditional convex-partitioning methods are extremely fast but ineffective at minimizing the number of guards, while traditional sampling methods, designed to balance runtime and solution quality, are outperformed by the proposed HAR methods, which leverage the strengths of both approaches.
Additionally, we showcase a promising adaptation of the best-performing HAR method to the localization-uncertainty visibility model, achieving the required coverage ratio for small to moderate localization uncertainty.

Future work could apply HAR to visibility-driven route planning, such as efficient mobile robot inspection or search in mapped environments.
Moreover, new research directions may explore more realistic localization uncertainty models and develop sensor-placement methods with formal coverage guarantees under these models, enabling their seamless integration into the HAR framework.
Another avenue for future research is to enhance HAR with a more sophisticated refinement step to further reduce the number of guards, potentially adapting \emph{minimum set cover problem} metaheuristics~\citep{Rosenbauer2020} for the SPP\@.

\section*{Declaration of generative AI and AI-assisted technologies in the writing process}

During the preparation of this work, the authors used ChatGPT by OpenAI to refine the language of the manuscript.
After using this tool, the authors reviewed and edited the content as needed and take full responsibility for the content of the publication.

\bibliographystyle{elsarticle-num-names}
\bibliography{main}

\end{document}